\documentclass{Liebert_Author_revised}
\newtheorem{definition}{Definition}
\usepackage{subcaption}
\usepackage{float}
\usepackage{xcolor}
\usepackage{multirow}
\usepackage{epstopdf}
\usepackage{graphicx}

\title{A Mapper algorithm with implicit intervals and its optimization} 

\author{Yuyang Tao $^{1}$, Shufei Ge $^{1\ast}$ \\
{$^{1}$ Institute of Mathematical Sciences ShanghaiTech University}\\
{393 Middle Huaxia Road, Shanghai, 201012, China}\\
{E-mail: geshf@shanghaitech.edu.cn}}

\begin{document} 

\maketitle

\keywords{Topological Data Analysis, Mapper Graph, Gaussian Mixture Model, Extended Persistence Homology, Stochastic gradient descent}

\begin{abstract}
The Mapper algorithm is an essential tool for visualizing complex, high dimensional data in \textcolor{black}{topological} data analysis (TDA) and has been widely used in biomedical research. It outputs a combinatorial graph whose structure \textcolor{black}{encodes} the shape of the data. \textcolor{black}{However,} the need for manual parameter tuning and fixed \textcolor{black}{(implicit)} intervals, along with fixed overlapping ratios may impede the performance of the standard Mapper algorithm. Variants of the standard Mapper algorithms have been developed to address these limitations, yet most of them still require manual tuning of parameters. Additionally, many of these variants, including the standard version found in the literature, were built within a deterministic framework and overlooked the uncertainty inherent in the data. To relax these limitations, in this work, we introduce a novel framework that implicitly represents intervals through a hidden assignment matrix, enabling automatic parameter optimization via stochastic gradient descent. In this work, we develop a soft Mapper framework based on a Gaussian mixture model(GMM) for flexible and implicit interval construction. We further illustrate the robustness of the soft Mapper algorithm by introducing the Mapper graph mode as a point estimation for the output graph. Moreover, a stochastic gradient descent algorithm with a specific topological loss function is proposed for optimizing parameters in the model. Both simulation and application studies demonstrate its effectiveness in capturing the underlying topological structures. In addition, the application to an RNA expression dataset obtained from the Mount Sinai/JJ Peters VA Medical Center Brain Bank (MSBB) successfully identifies a distinct subgroup of Alzheimer’s Disease. The implementation of our method is available at \url{https://github.com/FarmerTao/Implicit-interval-Mapper.git}.
\end{abstract}

\section{Introduction}\label{sec1}

The Mapper algorithm is a powerful tool in \textcolor{black}{topological} data analysis (TDA) used to explore the shape of datasets. Initially introduced for 3D object recognition \citep{TopologicalMethods3D}, the Mapper algorithm has demonstrated remarkable efficiency in extracting topological information across various data types. Its application spans various areas, notably in biomedical research \citep{skafTopologicalDataAnalysis2022}. For example, it successfully identified new breast cancer subgroups with superior survival rates \citep{nicolauTopologyBasedData2011}. In single cell RNA-Seq analysis, a Mapper-based algorithm was used to study unbiased temporal transcriptional regulation \citep{SinglecellTopologicalRNAseq2017}. The variants of Mapper algorithms were also used to uncover higher-order structures of complex phenomics data \citep{HyppoXScalableExploratory2021}. Moreover, many efforts have been made to integrate the Mapper algorithm with other machine learning techniques. For instance, \cite{bodnarDeepGraphMapper2021} enhanced the performance of a graph neural network by  incorporating the Mapper algorithm as a pooling operation. When combined with autoencoders, the Mapper algorithm can be used as a robust classifier \citep{MapperBasedClassifier2019a}, remedying the shortcomings of traditional convolutional neural networks, such as susceptibility to gradient-based attacks. Recently, a visualization method called ShapeVis which was inspired by the Mapper algorithm was proposed to produce more concise topological structure than standard Mapper graph \citep{ShapeVis}. Besides advances of the Mapper algorithms in applications, there has been some new progress in theory. \cite{carriereStatisticalAnalysis2018} analyzed the Mapper algorithm from statistical perspective and introduced a novel method for parameter selection. Other studies have explored the algorithm's convergence to Reeb graphs \citep{brownProbabilisticConvergenceStability2021}, and examined its  topological properties \citep{deyTopologicalAnalysisNerves2017}.

 The standard Mapper algorithm generates a graph that captures the shape of the data by dividing the support of the filtered data into several overlapped intervals with a fixed length \citep{TopologicalMethods3D}. \textcolor{black}{Two} major limitations of the standard Mapper algorithm are the requirements of fixed length of intervals and overlapping rate. Many algorithms were proposed to address these limitations. For instance, F-Mapper \citep{FMapper2020} utilized the fuzzy clustering algorithm to allow flexible interval partitioning by introducing an additional threshold parameter to control the degree of overlap. However, the introduction of the parameter also increased the algorithm's computational complexity. In a different vein, the ensemble version of Mapper was proposed by \cite{kangEnsembleMapper2021,EnsembleLearningMapper2023}, which can recover the topology of datasets without parameter tuning. However, this method has a significant high computational cost due to the execution of multiple instances of the Mapper algorithm. Differently, the Ball Mapper \citep{BallMapper2019} directly constructed cover on a given dataset, avoiding the difficulty of choosing filter functions. Nevertheless, its construction process is somewhat arbitrary and could inadvertently introduce extra parameters, increasing computational complexity. The D-Mapper, as introduced by \cite{tao2024distribution}, developed a probabilistic framework to construct intervals of the projected data based on a mixture distribution. This approach offers flexibility in interval construction. Additionally, the Soft Mapper, proposed by \cite{DifferentiableMapper2024}, a smoother version of the standard Mapper algorithm. This algorithm focused on the optimization of filter functions by minimizing the topological loss.
 
In this study, our objective is to enhance the Mapper algorithm by incorporating a probability model that requires fewer parameter selections and enables flexible interval partitioning. Inspired by the D-Mapper \citep{tao2024distribution}, our approach  utilizes a mixture distribution. Still, it implicitly represents intervals through a hidden assignment matrix, eliminating the need for the interval threshold $\alpha$. This framework optimizes the mixture distribution by integrating stochastic gradient descent, facilitating automatic parameter tuning. This research contributes three key innovations as follows. First, we introduce a novel framework for constructing Mapper graphs based on mixture models, enabling flexible and implicit interval definitions. Second, we consider the uncertainty in a Mapper graph and introduce a mode-based estimation for the Mapper graph. Finally, we develop a stochastic gradient descent algorithm to optimize these intervals, resulting in an improved Mapper graph.

In addition, the simulation results show that our method is comparable to the standard Mapper algorithm and is more robust. Our model is also applied to an RNA expression dataset obtained from the Mount Sinai/JJ Peters VA Medical Center Brain Bank (MSBB) \citep{MSBB} and successfully identifies a distinct subgroup of Alzheimer’s Disease.

\section{Preliminary}
\subsection{Mapper algorithm}
The Mapper algorithm has been widely used to output a graph of a given dataset to represent its topological structure \citep{deyTopologicalAnalysisNerves2017}, the output graph can be regarded as a discrete version of the Reeb graph. The core concept of the Mapper algorithm is as follows. First, it projects a given high-dimensional dataset into a lower-dimensional space. \textcolor{black}{Then, it constructs a reasonable cover for the projected data, pulls back the original data points according to each interval, and clusters the points within each preimage (or pullback set).} Finally, each cluster is treated as a node of the output graph, and an edge is added between two clusters if they share any data points.

\begin{figure}[h]%
\centering
\includegraphics[width=1\textwidth]{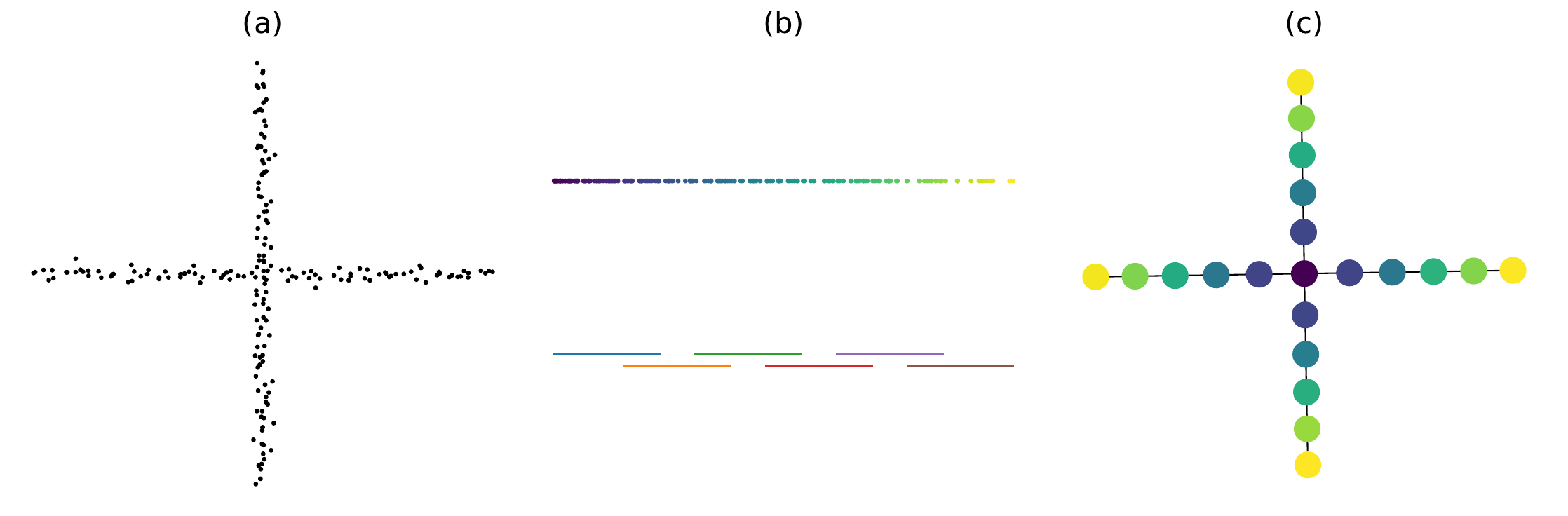}
\vspace{1mm}
\caption{A demonstration of the Mapper algorithm applied to a dataset with a cross structure. (a) A visualization of the dataset. (b) The projected data and its overlapped intervals when $K=6$, $p=0.33$. (c) The output graph of the Mapper algorithm. The clustering algorithm implemented here is the DBSCAN$(\epsilon = 0.6, minPts = 5)$ \citep{DBSCAN}. The output Mapper graph presents a cross shape, which is consistent with the shape of the dataset.}
\label{crosss_data}
\end{figure}

Consider a dataset with $n$ data points, $\{ \boldsymbol{x_1}, \ldots, \boldsymbol{x_n} \}$, $\boldsymbol{x_i} \in \mathbb{R}^d$, $i=1,\ldots, n$. Let $\mathbb{X}_n$ be the support of this dataset. Denote $f(\cdot): \mathbb{X}^n\rightarrow \mathbb{R}^m$, $m<n$,  a single-valued filter function, which maps each data point from $\mathbb{X}^n$ to a lower-dimensional space, $\mathbb{R}^m$ ($m\ge 1$). As in most of literature \cite{FMapper2020, kangEnsembleMapper2021,EnsembleLearningMapper2023}, here $m$ is set to $1$.  We use a simple example to illustrate the Mapper algorithm. The example dataset has a cross shape, as shown in Figure \ref{crosss_data} (a). The Mapper algorithm first projects data onto a real line using a filter function, $ f(\boldsymbol{x}) = y, ~\boldsymbol{x}\in \mathbb{R}^2,~ y \in \mathbb{R}$, as shown in the upper \textcolor{black}{panel} of Figure \ref{crosss_data} (b), the filter function is set to $f(\boldsymbol{x})=\frac{1}{n}\sum_{i=1}^{n}||\boldsymbol{x}-\boldsymbol{x}_i||$, i.e.  the mean distance of $\boldsymbol{x}$ to all points $\boldsymbol{x}_i$, $i=1,\ldots,n$. 
Set $a = \min \{f(\boldsymbol{x}_i), i=1,\ldots,n\}$, $b = \max \{f(\boldsymbol{x}_i), i=1,\ldots,n\}$. Then divide $[ a,b ]$ into $K$ equal length intervals with  $p$ percentage overlapping ratio between any two adjacent intervals, denoted as $I_j=[a_j,b_j]$, $j=1,\ldots,K$, as shown in the lower panel of Figure \ref{crosss_data} (b). These intervals such that $[ a,b ]= \bigcup_{j=1}^{K} I_j$, $|b_j-a_{j+1}|=p|b_1-a_1|$, $j=1,\ldots, n-1$, where $a=a_1<b_1<a_2<\ldots<a_n<b_{n-1}<b_n=b$.

 Each interval $I_j$ is then pulled back to the original space through the inverse mapping $f^{-1}(I_j), j= 1,...,K$. A clustering algorithm is applied to original data points fall into each $f^{-1}(I_j)$, partitioning the points into $r_j$ disjoint clusters, represented as $ C_{j,i}$, $i=1,\ldots,r_j$, such that $\bigcup_{r_j} C_{j,r_j} \subset f^{-1}(I_j)$ ,where $j= 1,...,K$. \textcolor{black}{We then obtain the pullback cover}, $C = \{C_{1,1},...,C_{1,r_1},...,C_{K,1},...,C_{K, r_K}  \}$. The Mapper graph is constructed from this cover, where each cluster in $C$ corresponds to a node in the graph. An edge is added between any two nodes if their corresponding clusters intersect, i.e., if $C_{i,j} \cap C_{i',j'} \neq \varnothing$. As shown in Figure \ref{crosss_data} (c), the resulting Mapper graph accurately captures the shape of the data. 

\subsection{Soft Mapper}

Traditional Mapper graphs are constructed through intervals based on the filtered data. The soft Mapper constructs Mapper graphs with a hidden assignment matrix without requiring fixed intervals~\citep{DifferentiableMapper2024}. The hidden assignment matrix\textcolor{black}{,} a ${n\times K}$ matrix, depicts the allocation relationship between given $n$ data points and $K$ groups \textcolor{black}{(i.e. the implicit $K$ intervals)}. This matrix is denoted as $H=(H_{ij})_{n\times K}$, in which $H_{ij} = 1$ if the $i$-th point belongs to $j$-th \textcolor{black}{group (implicit interval)}, otherwise $H_{ij} = 0$, for $i=1,\ldots,n,~j=1,\ldots,K$. With a hidden assignment matrix $H$, a Mapper graph is constructed directly through the standard process of pulling back and clustering, which we refer to as a Mapper function, defined as follows.

\begin{definition}[Mapper function] 
A Mapper function $\phi: H \mapsto G$, is a map from a hidden assignment matrix $H$ to a Mapper graph $G$. \textcolor{black}{The function is defined by pulling back and clustering operations in the standard Mapper algorithm.}
\end{definition}

When a hidden assignment matrix $H$ is a random matrix, the soft Mapper can be viewed as a stochastic version of the Mapper, parametrized by a Mapper function $\phi$ and a probability density function defined over a hidden assignment matrix $H$. The resulting random graph $G = \phi(H)$ is a function of a hidden assignment matrix $H$. The simplest example of a soft Mapper is obtained by assigning a Bernoulli distribution to each element $H_{ij}$ of a hidden assignment matrix $H$.

\begin{definition}[Soft Mapper with a Bernoulli distribution]
Suppose $H_{ij}$ follows a Bernoulli distribution with parameter $Q_{ij}$,
$$H_{ij} \sim \mathcal{B}(Q_{ij}),~i=1,\ldots,n, j=1,\ldots,K.$$
Where $Q = (Q_{ij})_{n \times K}$ is a probability matrix of the Bernoulli distributions. Each element of a hidden assignment matrix $H$ can be drawn independently from a Bernoulli distribution with probability of success $Q_{ij}$, $0\le O_{ij} \le 1$ for $i=1,\ldots,n,~j=1,\ldots,K$.
\end{definition}

With this definition, the model inference is simplified to estimate the probability matrix $Q$, which will yield the distribution of the Mapper graph. However, explicitly estimating $Q$ is challenging. In this work, we address this issue by making certain modifications to the soft Mapper approach.

\section{Methods}\label{sec2}

\subsection{GMM Soft Mapper}
\textcolor{black}{In the standard Mapper algorithm, a cover is constructed on the projected data, and a cover consists of several intervals and two adjacent intervals overlap. These intervals can be constructed by partitioning the projected data into multiple groups. Then, pull back the data points by allocating the corresponding projected value into each interval.  Analogous, this allocating process of projected data points into intervals of a cover is similar to the allocating process of labels for a mixture probability model. In this work, we focus on a Gaussian mixture model (GMM) to fit the projected data due to its simplicity and flexibility.} By incorporating a GMM, we can naturally define a hidden assignment matrix $H$ through soft clustering, and a probability matrix $Q$ can be easily derived from the GMM parameters. We fit a GMM to the projected data $\{ y_1,...,y_n \}$. As a result, the $Q_{ij}$ represents the probability that data point $y_i$ is assigned to $j$-th class within the GMM framework.

\begin{definition}[GMM Soft Mapper]
\label{GMM based Soft Mapper}

Assuming that the projected data $\{ y_1,...,y_n \}$ follows a Gaussian mixture distribution. We define the weights, means, and variances of each component as $\boldsymbol{\pi} = \{ \pi_1, \dots , \pi_K \}$, $\boldsymbol{\mu} = \{ \mu_1, \dots, \mu_K \}$, $\boldsymbol{\sigma^2} = \{ \sigma_1^2,\dots,\sigma_K^2 \}$, respectively. The set $\boldsymbol{\theta} = \{\boldsymbol{\pi}, \boldsymbol{\mu}, \boldsymbol{\sigma^2} \}$ denotes all model parameters. Conditional on $\boldsymbol{\theta}$, the distribution of each data point $y_i$ is given by:
$$y_i|\boldsymbol{\theta} \sim \sum_{k=1}^{K} \pi_k \mathcal{N}(y_i|\mu_k,\sigma_k^2).$$


Then, for each point $y_i$, it is natural to set the probability of $y_i$ belonging to $j$-th class as follows,

$$Q_{ij}(y_i) = \frac{\pi_j \mathcal{N}(y_i|\mu_j,\sigma_j^2)}{\sum_{k=1}^{K} \pi_k \mathcal{N}(y_i|\mu_k,\sigma_k^2)}.$$
\textcolor{black}{where $i = 1, \dots , n, j = 1, \dots, K$.}
\end{definition} 

This GMM-based assignment scheme imposes an implicit constraint on each row of the probability matrix $Q$, such that the sum of the probabilities for each data point equals one, $\sum_{j=1}^K Q_{ij}=1$. However, the independent Bernoulli distribution assignment scheme of $H_{ij}$ in the Soft Mapper may lead to extreme assignments, some points being unassigned to any clusters or a point is being assigned to all clusters. 
For instance, if $K=4$, data point $y_i$ might have an assignment probability vector of $Q_{i.}=[0.3,0.2,0.4,0.1]$, $i=1,\ldots,n$. Then the independent Bernoulli distribution assignment scheme may result an assignment vector $H_{i.}=[0,0,0,0]$, which means the data point $y_i$ is being unassigned to any clusters, or $H_{i.}=[1,1,1,1]$, which means the data point $y_i$ belongs to all clusters. The former case violates the Mapper algorithm requirement that each data point belongs to at least one cluster, and the extreme case of being unassigned to any clusters should be avoided. The latter case conflicts with that fact the a data point in the Mapper algorithms typically falls into, at most, two adjacent intervals.  Therefore, we propose a GMM-multinomial soft Mapper approach. Instead of sampling each element of $H$ \textcolor{black}{independently} through the Bernoulli distributions, we sample each row of the assignment matrix $H$ through a multinomial distributionconcerning each row of the probability matrix $Q$. The multinomial distribution grantees that each point is assigned to at least to one cluster, and at most to $m$ clusters as follows.

\begin{definition}[Soft Mapper with a multinomial distribution]
Let $Q_{i\cdot} = (Q_{i1},\ldots,Q_{iK})$ represent the $i$-th row column of $Q$, and $H_{i\cdot}= (H_{i1},\ldots,H_{iK})$ represent the $i$-th row column of hidden assignment matrix $H$, $i = 1, \dots , n, j = 1, \dots, K$. Assume each row $H_{i\cdot}$ follows a multinomial distribution with total number of events $m=2$ and event probability vector $Q_{i.}$, 
$$H_{i\cdot} \sim {Multi}(m,~Q_{i.}),~i=1,\ldots,n.$$

Here $H_{ij}$ may take values from $\{0,1,2\}$, and $H_{ij}\ge 1$ indicates that the $i$-th point is assigned to $j$-th \textcolor{black}{implicit interval (group)}, otherwise $H_{ij} = 0$. For ease of notations, we denote $H \sim {Multi}(m,~Q)$ to indicate that each row of $H$ independently follows a multinomial distribution with event number parameter $m=2$ and event probabilities being each row of $Q$.
\end{definition}

Our proposed soft Mapper samples each row $H_{i\cdot}$ independently from a multinomial distribution ${Multi}(2,Q_{i \cdot}),~i=1,\ldots,n$.  The total events $m$ of the multinomial distribution is set to $2$, guaranteeing that each data point \textcolor{black}{must be assigned to at least one group} and no more than two groups. This setting is reasonable, as in the Mapper algorithm, a data point typically falls within at least one group and at most two  groups.

Our approach differs from the soft Mapper described in \cite{DifferentiableMapper2024} in two key ways. Firstly, by using the GMM, our $Q_{ij}$ can be specified as a function of the parameters $\boldsymbol{\theta}$. Thus, updating the GMM parameters $\boldsymbol{\theta}$ is \textcolor{black}{equivalent to updating} $Q$. \textcolor{black}{In contrast, } $Q$ in \cite{DifferentiableMapper2024} was determined by the parameters in the filter functions. Secondly, we use an independent multinomial distribution assignment scheme to sample each row of the hidden assignment matrix $H$. The multinomial distribution avoids extreme cases where a data point is unassigned to any clusters or assigned to too many clusters.

\subsection{The Mapper graph mode}
The soft Mapper is a probabilistic version of the Mapper algorithm, where a random hidden assignment matrix $H$ can result in varying Mapper graphs. Although the topological structures of these soft Mapper graphs differ in detail, they often share some common features as they are derived from the same underlying distribution. We aim to obtain a Mapper graph that encapsulates all these common structures as its final representation. 
Since $H$ is a random discrete matrix, and each row is independently distributed, we can take the matrix of which each row is the mode of each row of $H$ as the mode representation of $H$ and then apply the Mapper function to get a Mapper graph, which we call the Mapper graph mode. The mode of a multinomial distribution has an explicit formula, allowing us to compute the mode of the Mapper graph directly and efficiently.

\begin{definition}[The mode of a soft Mapper with a multinomial distribution]
For a soft Mapper with a multinomial distribution, each row follows an independent multinomial distribution. Given the $i$-th row of $Q$, we define,
$$Q_{i}^{*} = \max \{Q_{i1},...,Q_{iK}\},$$
$$i^* = \mathop{\arg\max}\limits_{i1,...,iK} \{Q_{i1},...,Q_{iK}\},$$
$$Q_{i}^{**} = \max \{Q_{i1},...,Q_{iK}\} \setminus \{Q_{i}^{*} \},$$
$${i}^{**} = \mathop{\arg\max} \limits_ {\{i1,...,iK\} \setminus \{i^*\}} \{Q_{i1},...,Q_{iK}\} \setminus \{Q_{i}^{*} \}.$$
Here, the $Q_i^*$ and $Q_i^{**}$ are the largest and the second largest elements of $i$-th row of $Q$, respectively. ${i}^{*}$ and ${i}^{**}$ are their corresponding indices. For \textcolor{black}{$i=1,\ldots,n$}, the mode of the $i$-th row of the hidden assignment matrix is then determined as follows (see Appendix A for the theoretical derivation). 


If $\frac{1}{2}Q_i^* > Q_i^{**}$,

$$H_{mode}({i,j}) = 
\begin{cases} 
1 & \text{if } j = i^{*}, \\
0 & \text{else} .
\end{cases}$$

If $\frac{1}{2}Q_i^* \leq Q_i^{**}$,
$$H_{mode}({i,j}) = 
\begin{cases} 
1 & \text{if } j = i^{*}, i^{**} , \\
0 & \text{else} .
\end{cases}$$

The mode of Mapper graph can be defined by a Mapper function and the mode of hidden assignment matrix $H_{mode}$,
$$G_{mode} = \phi(H_{mode}).$$
\label{mode_mapper}
\end{definition}

The explicit form of the $H_{mode}$ significantly reduces the computational burden of the algorithm. In the following sections, we will adopt the mode of the Mapper graph as the representation of the graph and present several Mapper graph samples to illustrate the inherent uncertainty of these graphs.

\subsection{Loss function}
The probability matrix $Q$ is derived from a GMM fitted to the projected data. Therefore, optimizing the Mapper graph is equivalent to optimizing the parameter $\boldsymbol{\theta}$ of the GMM. However, the maximum likelihood estimation of parameter $\boldsymbol{\theta}$ only accounts for the distributions of the projected data, neglecting the topological information of the Mapper graph. To construct an appropriate Mapper graph, it is essential to consider both the likelihood of the projected data and the topology of data simultaneously. We design a loss function that incorporates both the likelihood of projected data and topological information of the Mapper graph. The log-likelihood of the projected data is given by 
$$\log L(\mathbb{Y}_n|\boldsymbol{\theta}) = \sum_{i=1}^{n} \log \sum_{k=1}^{K} \pi_k \mathcal{N}(y_i|\mu_k,\sigma_k^2).$$

We adopt a similar formula to \cite{DifferentiableMapper2024} to measure the topological information loss. The topological information of a Mapper graph can be encoded by the extended persistence diagram \citep{carriereStructureStabilityOneDimensional2018}. A significant advantage of the extended persistence diagram is its ability to capture the branches of a Mapper graph. The branches of a Mapper graph are crucial in downstream data analysis. We denote the extended persistence diagram as $D$, where $D = \{ (b_i, d_i) \mid i=1, \dots ,M\}$ \textcolor{black}{is a multiset containing $M$ points}. We use a function $Pers: G \mapsto D$ to represent the process of computing \textcolor{black}{the} extended persistence diagram for a given Mapper graph. 

To compute the extended persistence diagram, we need to define a filtration function for each node of a Mapper graph. Typically, the filtration function of a node on a Mapper graph is defined as the average of the filtered data within that node. However, in order to integrate the model prameters into the optimization framework, we introduce a novel node filtration function for computing the extended persistence diagram. For each node $c$ in a Mapper graph, the filtration function on this node is 
$$f_{M}(c) =  \frac{\sum_{y_i \in c} \log L(y_i|\boldsymbol{\theta})}{card(c)}.$$

This function calculates the averaged log-likelihood of the data points assigned to node $c$, where the number of points in node $c$ is denoted as $card(c)$. The edge filtration is the maximum value of the corresponding paired nodes. 

To represent the topological information on the extended persistence diagram, a function that maps the diagram to a real number is defined, denoted as $l: D \mapsto \mathbb{R}$.  This function is called the persistence specific function, which measures the topological information captured in the extended persistence diagram. There are various functions available for representing overall topological information, such as persistence landscapes \citep{perslandscapes2015} or computing the bottleneck distance from a target extended persistence diagram \citep{Bauer2024BottleneckDisEXPHD}. \textcolor{black}{Not all points on a persistence diagram are meaningful, some points may generated through noises in the dataset. Typically, points with short persistence are considered noises, while those with long persistence are regarded as meaningful signals. An effective diagram should primarily consist of signal points, reflecting the significant topological structures of the dataset.} \textcolor{black}{The Mapper graph intermediates between the data and the persistent homology, so that a \textit{good} Mapper
graph has not just (relatively) small noise but in fact \textit{few} features attributable to noise.} To get a more robust Mapper graph, in this work, we adopt the averaged persistence which measures the averaged persistence time of each point as the persistence-specific loss, \textcolor{black}{assume there are $M$ features concerning the Mapper graph mode, the topological loss is defined as:}
$$l(D) = \frac{1}{M}\sum_{i=1}^{M} |d_i - b_i|.$$

This value offers a reasonable summary of the extended persistence diagram. An effective diagram should predominantly feature signal points. \textcolor{black}{The detonator in this term is used to avoid encouraging spurious structures in the Mapper graph.}

The extended persistence diagram $D$ is \textcolor{black}{derived from a Mapper graph $G$, while $G$ is generated by} the Mapper function of the hidden assignment matrix $H$. For notation simplicity, we denote
\begin{align}
\nonumber
l(H) & = l \circ Pers \circ \phi(H)   \\
\nonumber
     & = l \circ Pers(G)  \\
\nonumber
     & = l(D) .
\end{align}
Since the hidden assignment matrix $H$ is random, the resulting $l(H)$ is also random. The conventional way to deal with random loss function is taking expectation $E(l(H))$. However, this method is computationally expensive when using Monte Carlo sampling. Instead, we use the persistence-specific loss of the Mapper graph mode defined in Definition \ref{mode_mapper} to represent the overall topological loss concerning $H$, denoted as $l(H_{mode})$. The mode of the Mapper graph can be computed directly without sampling, which reduces the computational cost significantly.

The total loss function of the GMM soft Mapper can be defined as a weighted average of the negative log-likelihood of the filtered data and the topological loss concerning $H$ 

\begin{align}
\nonumber
    Loss(\boldsymbol{\theta}|\mathbb{X}_n,\mathbb{Y}_n) &= -\lambda_1 \frac{\log L(\mathbb{Y}_n|\boldsymbol{\theta})}{n} - \lambda_2 l(H_{mode}). \\
    \nonumber
\end{align}

Here, the first term is the averaged log-likelihood, representing the information carried out by the filter data. By averaging, we ensure that the log-likelihood is comparable across datasets of varying sample sizes. The parameters $\lambda_1$ and $\lambda_2$ control the relative weights of the likelihood of the projected data and the topological loss of the Mapper graph. 

\subsection{Stochastic gradient descent parameter estimation}
In this study, we utilize stochastic gradient descent algorithm (SGD) \citep{SGD} to optimize the loss function defined above. Our goal is to find parameters $\boldsymbol{\theta}$ for a GMM that minimize the total loss function. The optimization process aims to improve the topological structure of a Mapper graph while ensuring high likelihood of the filtered data. Once the optimal parameters are determined, we can either sample from the GMM to create a soft Mapper graph or directly compute the Mapper graph mode. Notably, our \textcolor{black}{method} does not require the derivation of an explicit gradient expression. This simplification is facilitated by automatic differentiation frameworks such as PyTorch \citep{paszke2019pytorch} or TensorFlow \citep{abadi2016tensorflow}, which effectively implement SGD.

To apply SGD to the log-likelihood of a GMM, it is crucial to handle the constraints on the parameters. For weights $\pi$, each SGD update step must ensure that $\sum_{i=1}^{K} \pi_i = 1$, and $\pi_i > 0,i=1,\ldots,K$. To solve this problem, we adopt the method from \cite{GradientBasedTrainingGaussian2021}, which introduces free parameters $\xi_i, i=1,..,K$. Let 
$$\pi_i = \frac{e^{\xi_i}}{\sum_{i=1}^{K} e^{\xi_i}}, i=1, \dots ,K.$$

Instead of updating $\pi_i$ directly, this approach updates $\xi_i$. This transformation guarantees that $\pi_i$ meets the constraints at each step. Similarly, to satisfy the constraint of standard variance $\sigma_i> 0$, we take the transformation $\log \sigma_i = \xi'_i$. Then we update $\xi'_i$ instead of $\sigma_i$ at each step. The detailed SGD algorithm for the GMM soft Mapper model is presented in Algorithm \ref{SGD}. \textcolor{black}{In this work, we implement the proposed algorithm in Python version 3.10. The extended persistence diagram and bottleneck distance is computed by Python package GUDHI version 3.8.0, the stochastic gradient descent is implemented through pytorch 1.13.}

\begin{algorithm}[h!]
\SetAlgoLined

\KwIn{Learning rate $\gamma$, steps $N$, weights $\lambda_1,\lambda_2$, number of \textcolor{black}{implicit} intervals $K$, \\
initial parameters $\boldsymbol{\pi}^{(0)}, \boldsymbol{\mu}^{(0)}, \boldsymbol{\sigma^2}^{(0)}$.}
 \For{t = 1 \KwTo $N$}{
   Update $Q^{(t-1)}$ based on $\boldsymbol{\pi}^{(t-1)}, \boldsymbol{\mu}^{(t-1)}, \boldsymbol{\sigma^2}^{(t-1)}$; \\
   Compute the mode hidden assignment matrix $H_{mode}^{(t-1)}$; \\
   Compute the Mapper graph mode, $G_{mode}^{(t-1)} = \phi(H_{mode}^{(t-1)})$; \\
   Compute the persistence specific loss, $l(D_{mode}^{(t-1)})$; \\
   Compute the total loss, $Loss(\boldsymbol{\theta}^{(t-1)}|\mathbb{X}_n,\mathbb{Y}_n)$. \\
   \For{i = 1 \KwTo $K$}{
        $\xi_i^{(t)} = \xi_i^{(t-1)} - \gamma \frac{\partial Loss}{\partial \xi_i}|_{\xi_i^{t-1},\mu_i^{t-1},\xi^{\prime (t-1)}_i}$\;
        $\mu_i^{(t)} = \mu_i^{(t-1)} - \gamma \frac{\partial Loss}{\partial \mu_i}|_{\xi_i^{t},\mu_i^{t-1}, \xi^{\prime (t-1)}_i}$\;
        $\xi^{\prime (t)}_i = \xi^{\prime (t-1)}_i - \gamma \frac{\partial Loss}{\partial \xi'_i}|_{\xi_i^{t},\mu_i^{t},\xi^{\prime (t)}_i}$\;
        Compute weights, $\pi_i^{(t)} = \frac{e^{\xi_i^{(t)}}}{\sum_{i=1}^{K} e^{\xi_i^{(t)}}}$\;
        Compute variances, $\sigma_i^{(t)} = e^{\xi^{\prime (t)}_i }$.
   }
}
\caption{SGD algorithm for GMM soft Mapper}
\label{SGD}
\end{algorithm}

\section{Synthetic datasets}\label{experiments}
In this section, we compare our proposed algorithm to the standard Mapper algorithm \textcolor{black}{and D-Mapper} on synthetic datasets to demonstrate its effectiveness. The output graphs indicate that our proposed method \textcolor{black}{performs} comparable or better than the standard Mapper and D-Mapper algorithm concerning the topological structures. \textcolor{black}{To quantitatively compare these algorithms, we calculate the Silhouette Coefficient ($SC$) \citep{hanClusterAnalysis2012}} and the adjusted $SC$ ($SC_{adj}$) \citep{tao2024distribution} for each dataset in Section \ref{summary_sim}, see Table \ref{comparison}.
 In order to illustrate the uncertainty in the soft Mapper graph, we also output a few samples from the optimized distribution {(see Appendix B, figure \ref{Samples_two_cir} to figure \ref{Samples_3Dhuman}), and provide some summary statistics concerning some summary metrics of graphs in Table \ref{combined_graph_results} in Appendix B}. Training loss of the GMM soft Mapper while optimizing prameter for each dataset is given in Appendix E.

\subsection{Uniform dataset}
\begin{figure}%
\centering
\includegraphics[width=1.\textwidth]{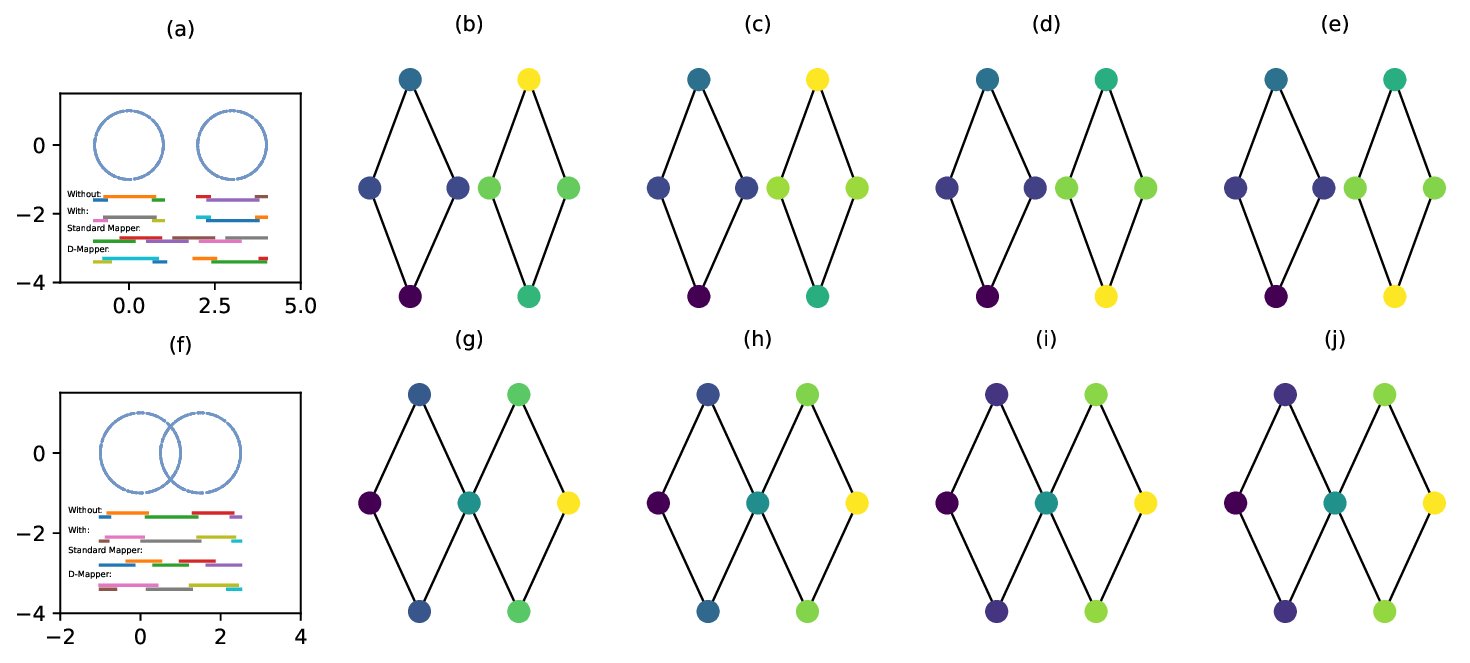}
\vspace{1mm}
\caption{Comparison of the standard Mapper algorithm, D-Mapper algorithm and our proposed algorithm on a two disjoint circles dataset and a two intersecting circles dataset. \textcolor{black}{(a, f). A visualization of the datasets. Coloured lines represents the \textcolor{black}{(implicit)} intervals produced by each algorithm. (b, g). The output graphs of the standard Mapper algorithm with $K=6, p=0.33$ for the two disjoint circles and $K=5, p=0.2$ for the two intersecting circles. (c, h). The output graphs of D-Mapper algorithm with $\alpha = 0.08$ for the disjoint circles and $\alpha = 0.31$ for the intersecting circles.  (d, i). The output Mapper graph mode of our proposed method without optimization. (e, j). The output Mapper graph mode of our proposed method with optimization.}}
\label{Mode_2c}
\end{figure}

We start with two uniformly sampled datasets. $1000$ data points are sampled from two separate circles and data sampled from two overlapping circles, respectively. Figures \ref{Mode_2c} (a) and (f) provide a visualization of these two datasets and \textcolor{black}{(implicit)} intervals produced by each algorithm. \textcolor{black}{To visualize the implicit intervals, we take the minimum point of each cluster as the starting point of the interval and the maximum point of each cluster as the ending point of the interval.} In both datasets, the filter function is the $x$-axis coordinates. \textcolor{black}{For the disjoint circles dataset, we set the number of \textcolor{black}{(implicit)} intervals to $K=6$ and apply the DBSCAN clustering algorithm with parameters $\epsilon = 0.3$ and $minPts = 5$. For the intersecting circles dataset, we use $K=5$ with the same clustering algorithm with parameters $ \epsilon = 0.2$ and $minPts = 5$.} Full details of the parameters are listed in Table \ref{tab1}. The Mapper graph modes, both with and without optimization, are depicted in Figures \ref{Mode_2c} (d), (i), and (e), (j). The \textcolor{black}{output graphs demonstrate that all the Mapper graphs effectively approximate the topological structures of these datasets}.

As the underlying data structure for these two small datasets is relatively simple, \textcolor{black}{all these three algorithms} can easily reveal the underlying structure. The topological structure of the data can even be well captured by the Mapper graph mode with an appropriate choice of the GMM initial values without optimization. 

\begin{figure}%
\centering
\includegraphics[width=1.\textwidth]{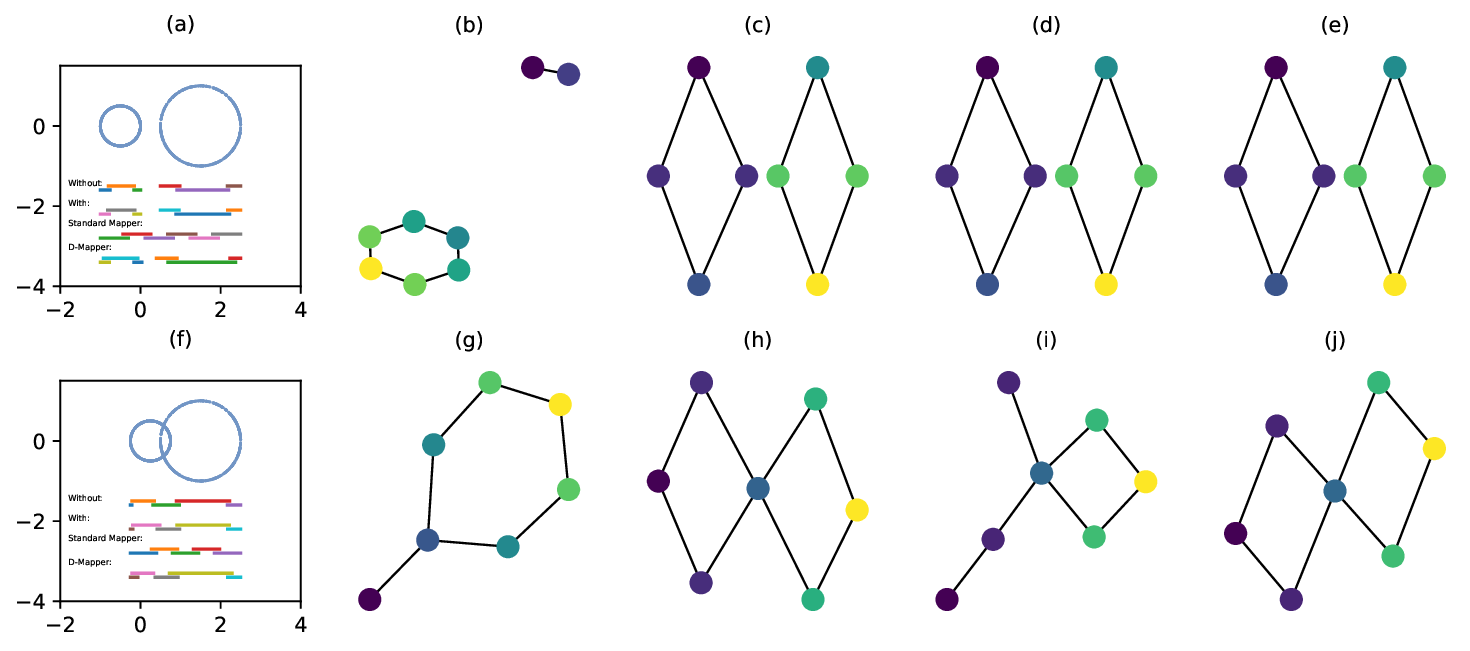}
\vspace{1mm}
\caption{Comparison of the standard Mapper algorithm, D-Mapper algorithm and our proposed algorithm on a two unequal-sized disjoint circles dataset and a two unequal-sized intersecting circles dataset. \textcolor{black}{(a, f). A visualization of the datasets. Coloured lines represents the \textcolor{black}{(implicit)} intervals produced by each algorithm. (b, g). The output graphs of the standard Mapper algorithm with $K=6, p=0.3$ for the disjoint circles and $K=5, p=0.2$ for the intersecting circles. (c, h). The output graphs of D-Mapper algorithm with $\alpha = 0.06$ for the disjoint circles and $\alpha = 0.05$ for the intersecting circles. (d, i). The output Mapper graph mode of our proposed method without optimization. (e, j). The output Mapper graph mode of our proposed method with optimization.}}
\label{Mode_2uc}
\end{figure}

We make slight modifications to the datasets above, resulting in two disjoint and two intersecting \textcolor{black}{circles with different radii}, each containing $1000$ points. The filter function remains the $x$-axis coordinates. For the unequal-sized disjoint circles, DBSCAN parameters are set to \textcolor{black}{$\epsilon = 0.35$ and $minPts = 5$,} for the unequal-sized intersecting circles, parameters are $\epsilon = 0.2$ and $minPts = 5$. \textcolor{black}{The results of the three algorithms are shown in Figure \ref{Mode_2uc}.} The standard Mapper struggles to capture the correct topological structure due to its fixed interval setting, whereas \textcolor{black}{the D-Mapper and }our method can adjust intervals based on data distribution, facilitating the accurate capture of the data's underlying shape. In the case of the unequal-sized intersecting circles dataset, the Mapper graph mode fails to correctly capture the data structure without optimization. However, after optimization, the \textcolor{black}{implicit intervals (groups)} are slightly adjusted, leading to more accurate results.

\subsection{Noisy dataset}
\begin{figure}%
\centering
\includegraphics[width=1.\textwidth]{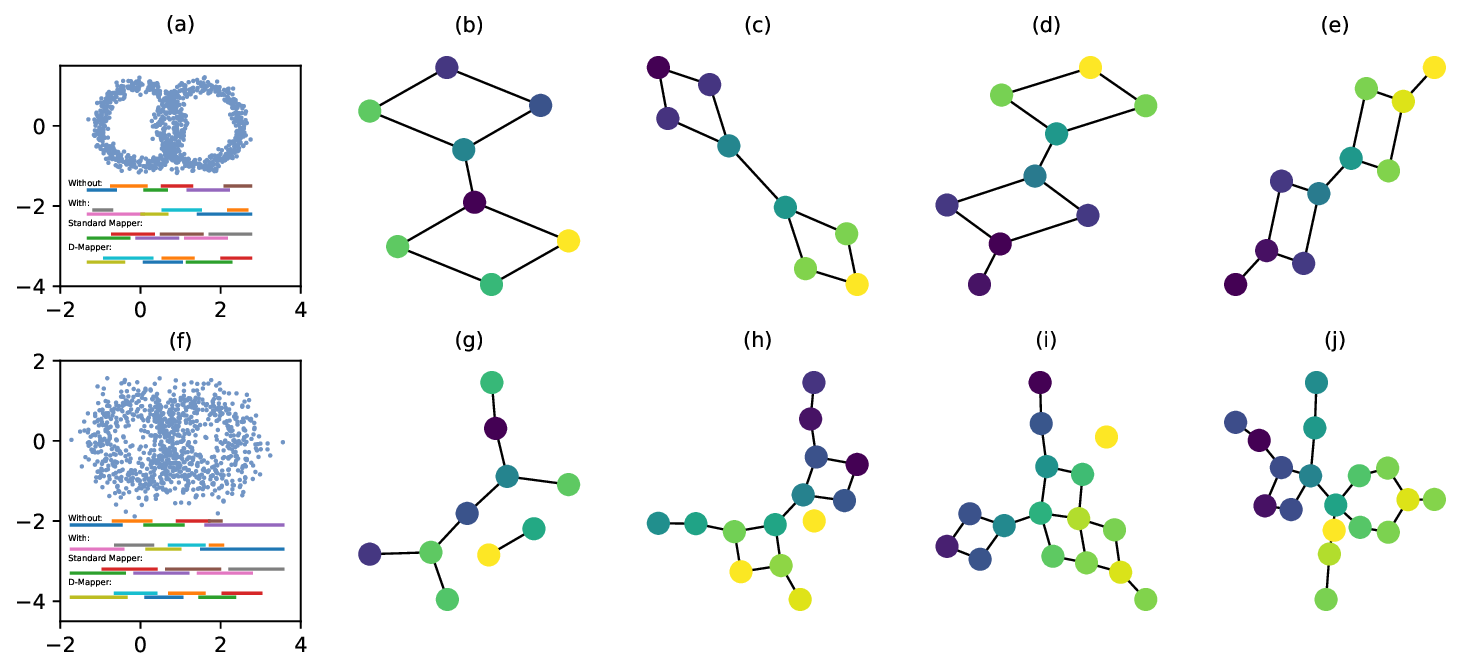}
\vspace{1mm}
\caption{Comparison of the standard Mapper algorithm, D-Mapper algorithm and our proposed algorithm on a two intersecting circles dataset with noises. \textcolor{black}{(a, f). A visualization of the datasets. Coloured lines represents the \textcolor{black}{(implicit)} intervals produced by each algorithm. (b, g). The output graphs of the standard Mapper algorithm with $K=6, p=0.4$ for the small noise dataset and $K=6, p=0.4$ for the big noise dataset. (c, h). The output graphs of D-Mapper algorithm with $\alpha = 0.025$ for the small noise dataset and $\alpha = 0.07$ for the big noise dataset. (d, i). The output Mapper graph mode of our proposed method without optimization. (e, j). The output Mapper graph mode of our proposed method with optimization.}}
\label{Mode_two_insert_cir_noise}
\end{figure}

In this section, we assess the robustness of our proposed algorithm by introducing noises into the intersecting circles dataset. The noise is drawn from a Gaussian distribution and is added to each point's coordinates. Initially, we introduce a small amount of noise, characterized by a Gaussian distribution with mean of zero and standard deviation of $0.1$. The dataset is shown in Figure \ref{Mode_two_insert_cir_noise} (a). With this setting for noises, both the standard Mapper, \textcolor{black}{D-Mapper and our Mapper graph mode can accurately represent the correct topological structure. The results are shown in Figure \ref{Mode_two_insert_cir_noise} (b)-(e).}

Subsequently, we increase the noise level, increasing the standard deviation up to $0.3$. The dataset is shown in Figure \ref{Mode_two_insert_cir_noise} (f). At this noise level, the standard Mapper algorithm is unable to output meaningful structures, as illustrated in Figure \ref{Mode_two_insert_cir_noise} (g). \textcolor{black}{The D-Mapper algorithm can captuare the main two loops of the data, while still has some extra branches and an isolated node, as shown in Figure \ref{Mode_two_insert_cir_noise} (h).} After optimization, our Mapper graph mode still can capture the two primary loops, as shown in Figure \ref{Mode_two_insert_cir_noise} (j), though with some additional branches. These results highlight the efficacy of flexible \textcolor{black}{implicit} interval partitioning in uncovering the data's inherent structure, even in the presence of significant noise. The parameter optimization plays a crucial role in accurately determining the \textcolor{black}{implicit} intervals, which boosts the algorithm's performance.

\subsection{3D human dataset}
\begin{figure}
\centering
\includegraphics[width=1.\textwidth]{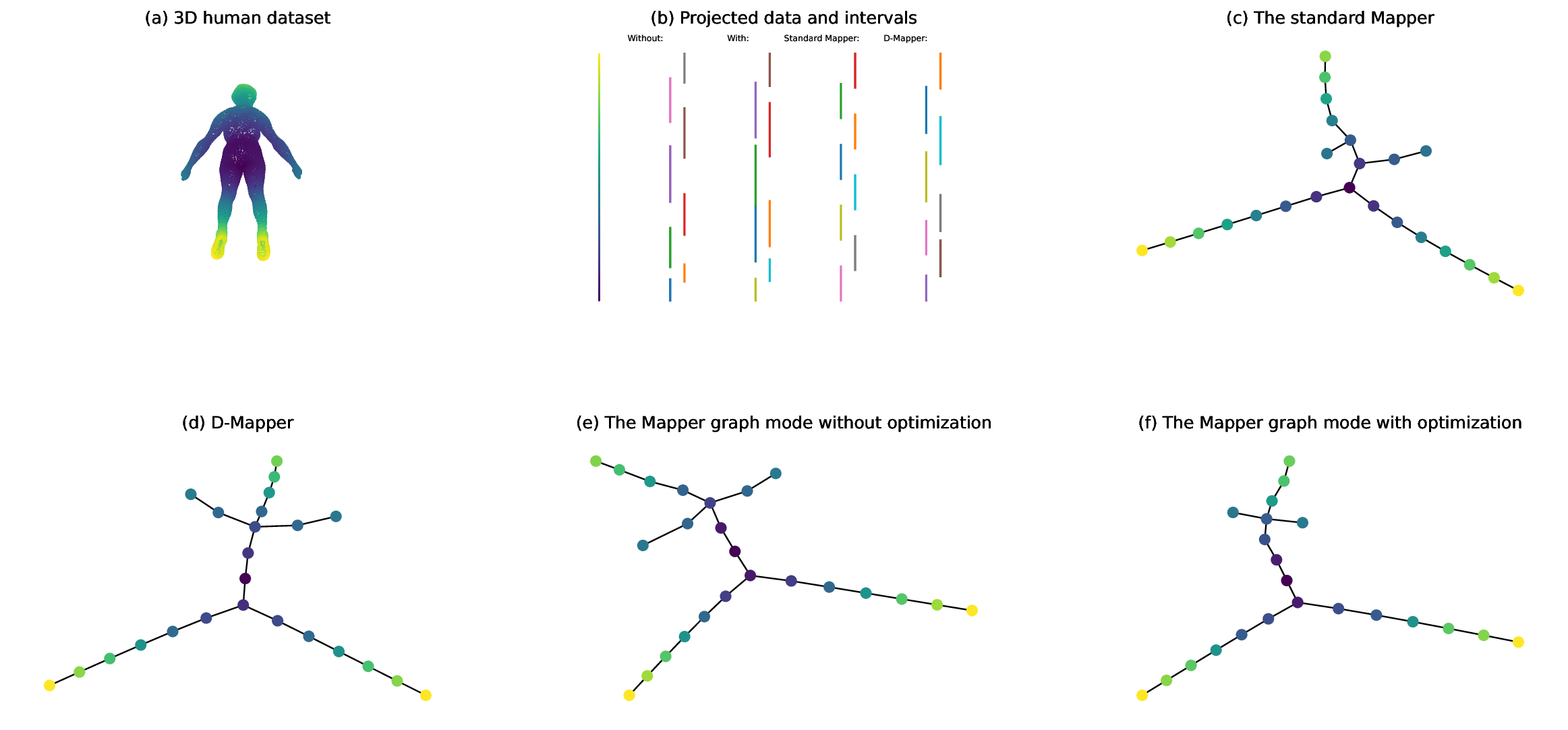}
\caption{Comparison of the standard Mapper algorithm, D-Mapper and our proposed algorithm on a 3D human dataset. \textcolor{black}{(a) A visualization of the 3D human dataset. (b) The projected data and resulted \textcolor{black}{(implicit)} intervals of each algorithm. (c) The output graph of the standard Mapper algorithm, the number of intervals is $8$ and the overlap rate is $0.1$. (d) The output graph of D-Mapper algorithm, the parameter $\alpha = 0.071$. (e) The Mapper graph mode without optimization. (f) The Mapper graph mode with optimization.} }
\label{3Dhuman}
\end{figure}

We finally test our proposed algorithm on a 3D human dataset from \cite{DifferentiableMapper2024}. In this example, the number of \textcolor{black}{(implicit)} intervals is set to $K=8$ and the DBSCAN clustering with parameters \textcolor{black}{$\epsilon = 0.1, minPts = 5$} is used. The filter function is the mean value of distance between each sample and others. \textcolor{black}{The standard Mapper outputs an asymmetrical skeleton, while both the D-Mapper and the Mapper graph mode without optimization produce a symmetrical skeleton. After optimization, the Mapper graph mode outputs a more concise structure (a symmetric skeleton with less nodes) than the standard Mapper graph and D-Mapper graph as shown in Figure \ref{3Dhuman}.} These results indicates that the proposed soft Mapper algorithm performs better than the standard Mapper concerning the visualization of topological structures, and the optimization process successfully further optimize the topological structures.

\subsection{Quantitatively summary of synthetic datasets}\label{summary_sim}

\textcolor{black}{Through a series of synthetic dataset experiments, we find that the optimized Mapper graph mode  can capture dataset shapes more accurately in general. 
It should be noted that the quantitative evaluation of the Mapper graphs is a challenging problem, as the topological information of a given data is often not available \citep{FMapper2020, carriereStatisticalAnalysis2018}. As in most of the literature,  here we compute the $SC$, this metric accounts for the clustering performance only \citep{FMapper2020, chalapathiAdaptiveCoversMapper2021, BallMapper2019}.  The $SC_{adj}$ accounts for both the clustering performance as well as the topological information, to some extent \citep{tao2024distribution}.}

\textcolor{black}{As shown in Table  \ref{comparison}, in all cases but the two intersecting dataset and the 3D human dataset, the D-Mapper performs comparably or better than the than the standard Mapper and the Mapper graph mode performs best in terms of metric $SC_{adj}$. For the two intersecting circles dataset, the $SC_{adj}$  of the D-Mapper is lower than the standard Mapper graph. This is mainly caused by the lower $TSR$ value of the D-Mapper. This dataset indicates that the D-Mapper algorithm may be unstable when the number of intervals $K$ ($=5$) is small for some cases, and when $K$ is set to a large number ($K=8$), denoted as D-Mapper* in Table  \ref{comparison}, a larger $TSR$ is obtained. The graph of D-Mapper* and more explanations can be found in Appendix C. For the 3D human dataset, the $SC_{adj}$ of the Mapper graph mode is a little bit lower than that of the standard Mapper and D-Mapper, but as the output graphs indicates (see Figure \ref{3Dhuman}), the Mapper graph mode has more condensed structures.}

\begin{table}
\begin{threeparttable}
\caption{Model parameters setting for synthetic datasets}
\label{tab1}
\begin{tabular}{lllll}
\toprule
Dataset & $K$  & clustering & learning rate & N  \\
\midrule
Two disjoint circles                & 6   & DBSCAN(0.3,5)  & 0.005  &200\\
Two intersecting circles      & 5   & DBSCAN(0.2,5)  & 0.01  &300\\
Two unequal disjoint circles        & 5   & DBSCAN(0.35,5)  & 0.001  &300\\
Two unequal intersecting circles    & 5   & DBSCAN(0.2,5)  & 0.001  &400\\
Two intersecting circles with small noise & 6   & DBSCAN(0.2,5)  & 0.002 &250\\
Two intersecting circles with big noise & 6   & DBSCAN(0.2,5)  & 0.001 &300\\
3D human                            & 8   & DBSCAN(0.1,5)  & 0.0001 &200\\
\bottomrule
\end{tabular}
\end{threeparttable}
\end{table}

\begin{table}[ht]
\caption{Quantitative comparison of the Mapper, D-Mapper, and Mapper graph mode across different datasets. Generally, the D-Mapper has a higher value of metric $SC_{adj}$ than the standard Mapper and the Mapper graph mode performs best concerning metric $SC_{adj}$. The D-Mapper* refer to the D-Mapper but with a larger \textcolor{black}{number of intervals ($K=8$)}.}
\label{comparison}
\centering
\small
\begin{tabular}{l l c c c}
\toprule
\textbf{Datasets} & \textbf{Methods} & \textbf{SC\_norm} & \textbf{TSR} & \textbf{SC\_adj} \\
\midrule
\multirow{3}{*}{Two disjoint circles} 
    & Mapper & 0.59 & 1 & 0.8 \\
    & D-Mapper & 0.68 & 1 & 0.84 \\
    & Mapper mode & 0.73 & 1 & 0.86 \\
\midrule
\multirow{3}{*}{Two intersecting circles} 
    & Mapper & 0.58 & 1 & 0.79 \\
    & D-Mapper & 0.59 & 0.33 & 0.46 \\
    &D-Mapper* & 0.59 & 1 & 0.79 \\
    & Mapper mode & 0.65 & 1 & 0.83 \\
\midrule
\multirow{3}{*}{3D human} 
    & Mapper & 0.51 & 1 & 0.76 \\
    & D-Mapper & 0.5 & 1 & 0.75 \\
    & Mapper mode & 0.44 & 1 & 0.72 \\
\midrule
\multirow{3}{*}{Two unequal-sized disjoint circles} 
    & Mapper & 0.61 & 1 & 0.8 \\
    & D-Mapper & 0.63 & 1 & 0.82 \\
    & Mapper mode & 0.7 & 1 & 0.85 \\
\midrule
\multirow{3}{*}{Two unequal-sized intersecting circles} 
    & Mapper & 0.64 & 1 & 0.82 \\
    & D-Mapper & 0.63 & 1 & 0.82 \\
    & Mapper mode & 0.68 & 1 & 0.84 \\
\midrule
\multirow{3}{*}{Two intersecting circles with small noises} 
    & Mapper & 0.52 & 0.29 & 0.40 \\
    & D-Mapper & 0.55 & 0.33 & 0.44 \\
    & Mapper mode & 0.5 & 1 & 0.77 \\
\midrule
\multirow{3}{*}{Two intersecting circles  with big noises} 
    & Mapper & 0.43 & 0.25 & 0.34 \\
    & D-Mapper & 0.47 & 0.25 & 0.36 \\
    & Mapper mode & 0.6 & 0.48 & 0.54 \\
\bottomrule
\end{tabular}
\end{table}

\section{Application}
We also apply our method to an RNA expression dataset to assess its ability to identify subgroups within Alzheimer's disease (AD) patients. The dataset is obtained from the Mount Sinai/JJ Peters VA Medical Center Brain Bank (MSBB) \citep{MSBB}, which includes RNA gene expression profiles from four distinct brain regions, the frontal pole (FP) in Brodmann area 10, the superior temporal gyrus (STG) in area 22, the parahippocampal gyrus (PHG) in area 36, and the inferior frontal gyrus (IFG) in area 44. In this application study, we focus on brain area 36, which includes $215$ patient samples, each with over $20,000$ gene expression values. Each patient is given a Braak AD staging score, ranging from $0$ to $6$, with higher scores indicating more severe disease stages \citep{braak2003staging}. The Braak score is treated as a label for each sample. Our goal is to identify subgroups with a significantly different distribution of Braak scores from the rest of the population given the gene expression profiles. These subgroups can be  branches or isolated nodes on the hidden topological features of the gene expression profiles.

As indicated in \cite{Zhou2021Hyperbolic}, the gene expression data could exhibit a hyperbolic structure especially as the number of \textcolor{black}{genetic sites increases}.  To capture the complex and hierarchical relationships between samples based on gene expression more effectively, we use the Lorentzian distance, a standard hyperbolic distance, to measure the similarity between samples. The hyperbolic distance can preserve the hierarchical structures inherent in gene expression data and is considered particularly suitable for high-dimensional data \citep{liu2024bayesian}. In the Lorentzian distance, the centroid is crucial due to its ability to reveal the hierarchical structure of datasets. In this study, we set each data point in the dataset as the centroid to get a distance matrix repetitively and use the averaged distance matrix as the final distance matrix. The filter function is set to the mean value of distance between each sample and others. Due to the high computational cost of calculating the distance matrix for over $20,000$ genes, we select the top $37$ gene sites as listed in \cite{wang2016integrative}.  These gene sites are from the top $50$ probes ranked in association with disease traits across $19$ brain regions, excluding those unmatched. The selected gene sites are then used to compute the hyperbolic distances.

We set the number of \textcolor{black}{(implicit)} intervals to $K = 15$, the learning rate to $\gamma = 0.001$, and the number of epochs to $N = 300$. We apply agglomerative clustering with a threshold of $3.13$ \citep{agglomerative}. Additionally, we initialize the parameters by fitting a Gaussian mixture model to the projected data. The graph modes of the GMM soft Mapper algorihtm without and with optimation is given in Appendix D. Training loss of the GMM soft Mapper while optimizing prameter is given in the Appendix E. The resulting optimized Mapper graph mode offers valuable insights into Alzheimer's disease progression. The $\chi^2$ test is used to determine if there is a statistically significant difference between the distribution of the identified subgroup and the rest of the population \citep{mannan1984bird}.

\begin{figure}%
\centering
\includegraphics[width=\textwidth]{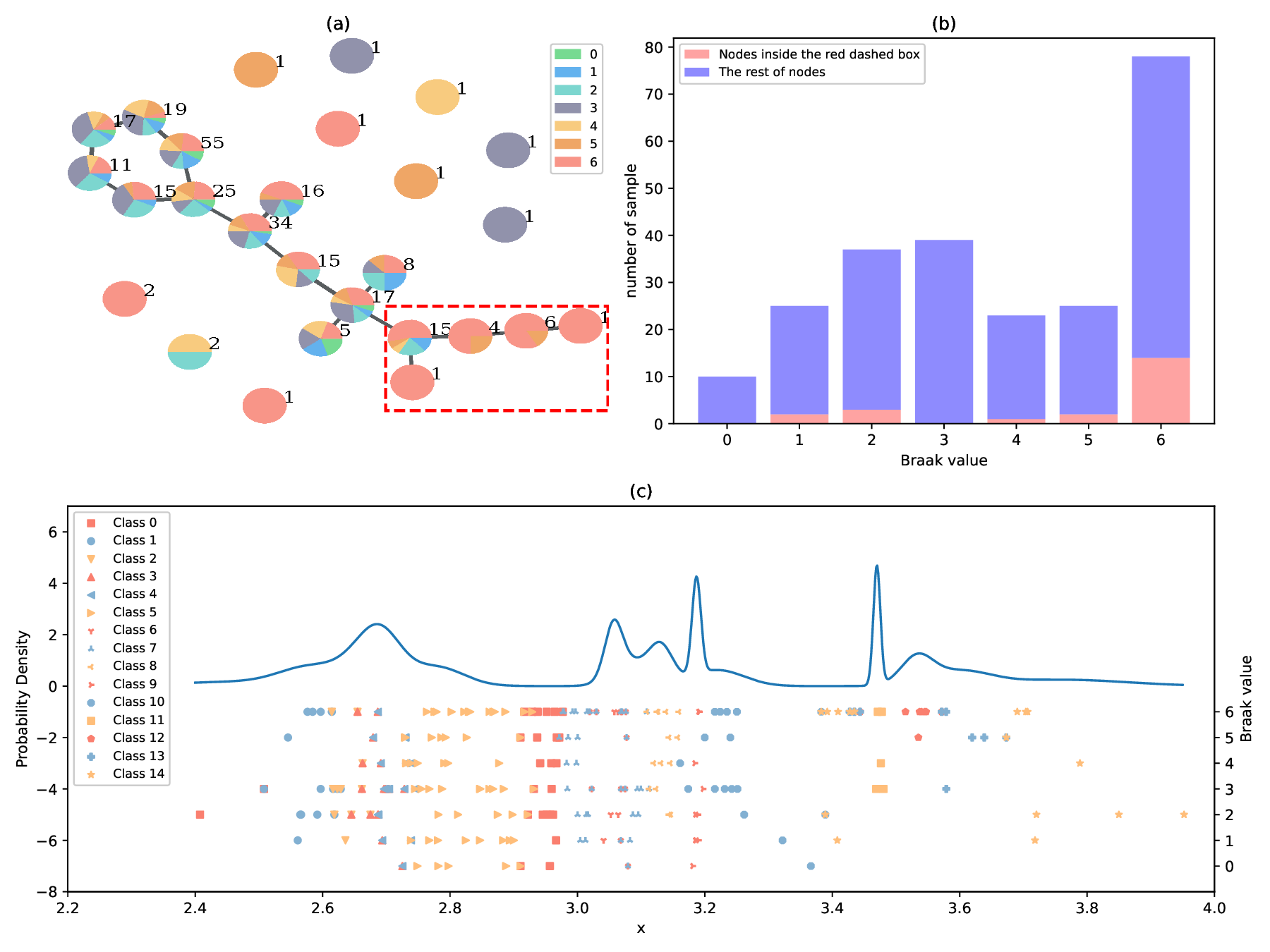}
\vspace{1mm}
\caption{Results of the brain area 36. (a) The output Mapper graph mode of a GMM soft Mapper with optimization. The pie charts at each node illustrate the distribution of Braak scores within this node, with the number next to each node representing the sample size in the node. Nodes within the red dashed box is a cluster that exhibits a distinct distribution compared to samples in the rest nodes. (b) The bar chart compares the distribution of Braak scores of the nodes within the red dashed box to those of the rest nodes. (c) A visualization of the projected data along with the optimized GMM probability density. In this figure, the values of the major $y$-axis represent the GMM density and the values of the minor $y$-axis represent the Braak scores of each data point. Coloured markers signify the various implicit intervals to which a point belongs.}
\label{36}
\end{figure}

In brain area 36, we find a distinct branch which is associated with high Braak scores, as shown in the red dashed box in  Figure \ref{36} (a). The bar charts in Figure \ref{36} (b) reveals distribution differences between this branch and the \textcolor{black}{other} nodes. The $P-$value of the $\chi^2$ test is $0.0047$, less than $0.05$, indicating a significant difference in distribution. $73\%$ of patients in this unique group have severe Alzheimer's disease (with Braak scores $=5$ or $=6$),  while the averaged ratio in the rest nodes is $36.8\%$. This indicates that the gene expression patterns in this cohort differ from the rest, and merit further investigation. \textcolor{black}{We also run the standard Mapper and D-Mapper on this dataset and the results are shown in the Figure \ref{MSBB_compare} in Appendix D. The standard Mapper exhibits a distinct branch (in the red dashed box) that demonstrates a significant distributional difference compared to other nodes ({$81\%$ severe conditions  versus $37.1\%$ severe conditions with the $P-$value of the $\chi^2$ test is $0.0024$}), as indicated in Figure ~\ref{MSBB_compare} (a). However, this branch has more nodes and sub-branches, which indicates a more separate structure, than that of the Mapper graph mode. No similar branch is observed in the D-Mapper graph and it contains many scattered nodes (or connected components). This also further supports that one of the advantages of our algorithm is its ability to produce a more concise topological structure of data.}

\section{Discussion}\label{sec4}

In this work, we introduce a novel approach for flexibly constructing a Mapper graph based on a probability model.  We develop implicit intervals for soft Mapper graphs using a Gaussian mixture model and a multinomial distribution. With a given number of \textcolor{black}{implicit} intervals, our algorithm automatically assigns each data point to an \textcolor{black}{implicit} interval based on the allocation probability. Then based on these implicit intervals, we derive the concept of the Mapper graph mode as a point estimation. Additionally, we design an optimization approach that enhances the topological structure of Mapper graphs by minimizing a specific loss function. This function considers both the likelihood of the projected data with respect to the GMM and the topological information of the Mapper graph mode. This optimization process is particularly suited for complex and noisy datasets, as the standard Mapper algorithm can be sensitive to noise and may fail in certain situations.  Both simulation and application studies demonstrate its effectiveness in capturing the underlying topological structures. The application of our proposed algorithm to gene expression of brain area 36 from the MSBB successfully identifies a unique subgroup whose distribution of Braak scores significantly differs from the rest of the nodes.

It is worth noting that the multimodal nature of the objective function makes it challenging to find a globally optimal Mapper graph. \textcolor{black}{We suggest performing multiple optimization runs with careful parameter tuning and incorporating domain knowledge to generate high-quality Mapper graphs.} In addtion, the mean persistence loss function used in this work is a simple representation of information on the extended persistence diagram. Numerous alternative approaches may be available, for instance, one could consider distinguishing signals from noise in a Mapper graph using confidence sets \citep{2013Confidence}. Finding a more robust persistence-specific loss function will be a one of the \textcolor{black}{future} work directions. \textcolor{black}{Moreover}, as you may have noticed, the filter function is also important for constructing \textcolor{black}{a meaningful Mapper graph. In this work,} covers are constructed based on the filtered data, \textcolor{black}{as in most literature}. Optimizing the filter function and \textcolor{black}{(implicit)} intervals simultaneously would be a challenging but meaningful direction. \textcolor{black}{Moreover, a more appropriate quantitative assessment of the Mapper algorithms poses significant challenges. We adopt $SC_{adj}$ in this work, but this measure still has several limitations, for example, how to balance the trade-off between the clustering performance and the topological information. Besides, the favour of fine clustering of $SC$ may also limit the performance of the $SC_{adj}$. Developing a more robust and objective quantitative evaluation metric for Mapper algorithms constitutes a critical open problem in the field.
Lastly, how to use a Mapper algorithm to achieve association studies between genetic sites and phenotypes simultaneously is an insightful direction, for example, which will further empower the applications of the Mapper algorithms in computational biology.}

\section*{Acknowledgments}
This work was supported by the National Natural Science Foundation of China (12401383), the Shanghai Science and Technology Program (No. 21010502500),  the startup fund of ShanghaiTech University, and the HPC Platform of ShanghaiTech University.

\bibliographystyle{plainnat-revised}
\bibliography{main}

\newpage
\appendix
\section*{Appendix A. Theoretical derivation of the mode of a multinomial distribution}\label{secA1}

Suppose $(h_{1},..,h_{K})$ follows a multinomial distribution with total event number $m=2$,
$$(h_{1},..,h_{K}) \sim {Multi}(2,q_{1},..,q_{K}).$$
Then the probability density function of $(h_{1},..,h_{K})$ is 
$$p(h_{1},..,h_{K}) = \frac{2}{h_1!h_2!...h_n!} q_1^{h_1}q_2^{h_2}...q_K^{h_K}.$$

To compute the mode of a \textcolor{black}{multinomial} distribution, i.e., 
$$ \mathop{\arg\max}\limits_{(h_{1},..,h_{K})} p(h_{1},..,h_{K}) = \mathop{\arg\max}\limits_{(h_{1},..,h_{K})} \frac{2}{h_1!h_2!...h_n!} q_1^{h_1}q_2^{h_2}...q_K^{h_K}.$$

When the total event number $m=2$, the possible values  of $(h_{1},..,h_{K})$ can be divided into two following cases.\\

\noindent \textbf{Case 1.  Two elements of $(h_{1},..,h_{K})$ are $1$ and others are $0$. }\\

In this case, to get the maximum probability, 
denote $q^* = max(q_1,q_2,...q_K)$ the largest probability and $q^{**} = max\{ q_1,q_2,...q_K \} \setminus \{ q^* \}$ the second largest probability among $\{ q_1,q_2,...q_K \}$. $i^*$ and $i^{**}$ represent the index of $q^*$ and $q^{**}$. Therefore, we have $h_{i^*}=h_{i^{**}}=1$ and others to zero. The probability of this case is
$$\mathop{\arg\max}\limits_{(h_{1},..,h_{K})} \frac{2}{h_1!h_2!...h_n!} q_1^{h_1}q_2^{h_2}...q_K^{h_K} = 2q^*q^{**}.$$

\noindent \textbf{Case 2.  One element of $(h_{1},..,h_{K})$ is $2$ and others are $0$. }\\

In this case, we have $h_{i^*}=2$ and others to zero. The probability of this case is
$$\mathop{\arg\max}\limits_{(h_{1},..,h_{K})} \frac{2}{h_1!h_2!...h_n!} q_1^{h_1}q_2^{h_2}...q_K^{h_K} = (q^*)^2.$$

Therefore, to get the mode of a multinomial distribution with $m=2$, we only need to consider the following two  situations.

If $\frac{1}{2}q^* > q^{**}$,

$$h_{mode}(i) = 
\begin{cases} 
2 & \text{if } i = i^{*}, \\
0 & \text{otherwise}. 
\end{cases}$$

If $\frac{1}{2} q^* \leq q^{**}$,
$$h_{mode}(i) = 
\begin{cases} 
1 & \text{if } i = i^{*}, i^{**}, \\
0 & \text{otherwise}.
\end{cases}$$

\section*{Appendix B. Samples from the optimized GMM soft Mapper of synthetic datasets}

\begin{figure}[H]
\centering
\includegraphics[width=0.9\textwidth]{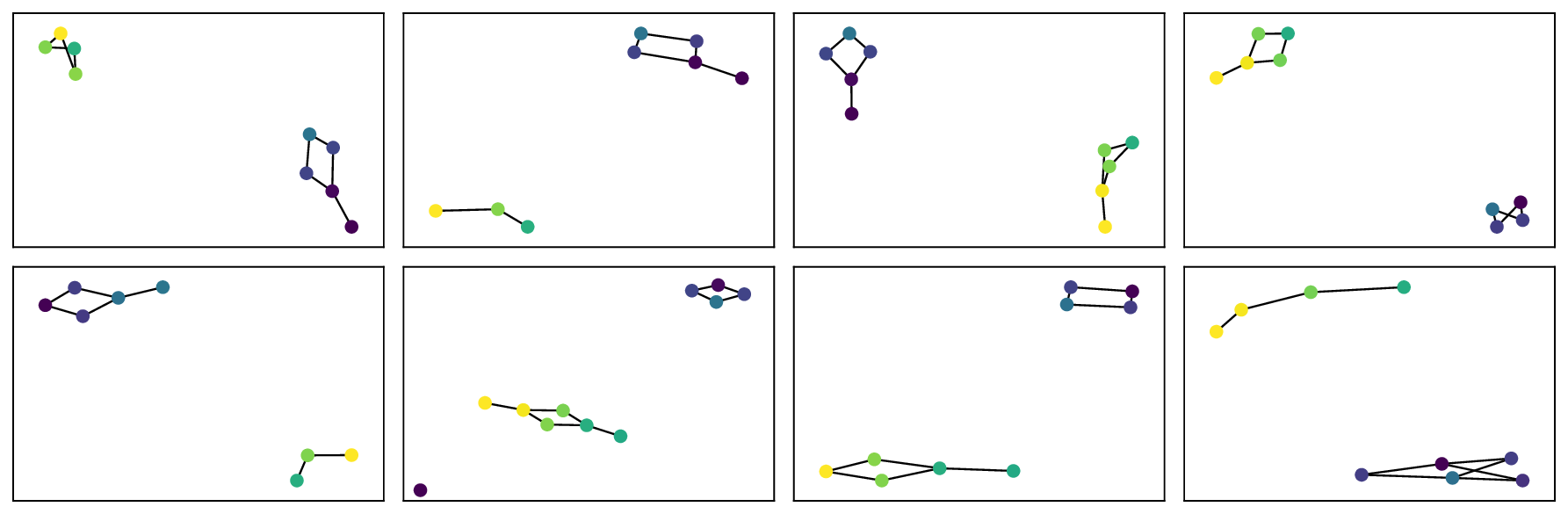}
\vspace{1mm}
\caption{Eight samples from the optimized GMM soft Mapper of the two disjoint circles dataset.}
\label{Samples_two_cir}
\end{figure}

\begin{figure}[H]
\centering
\includegraphics[width=0.9\textwidth]{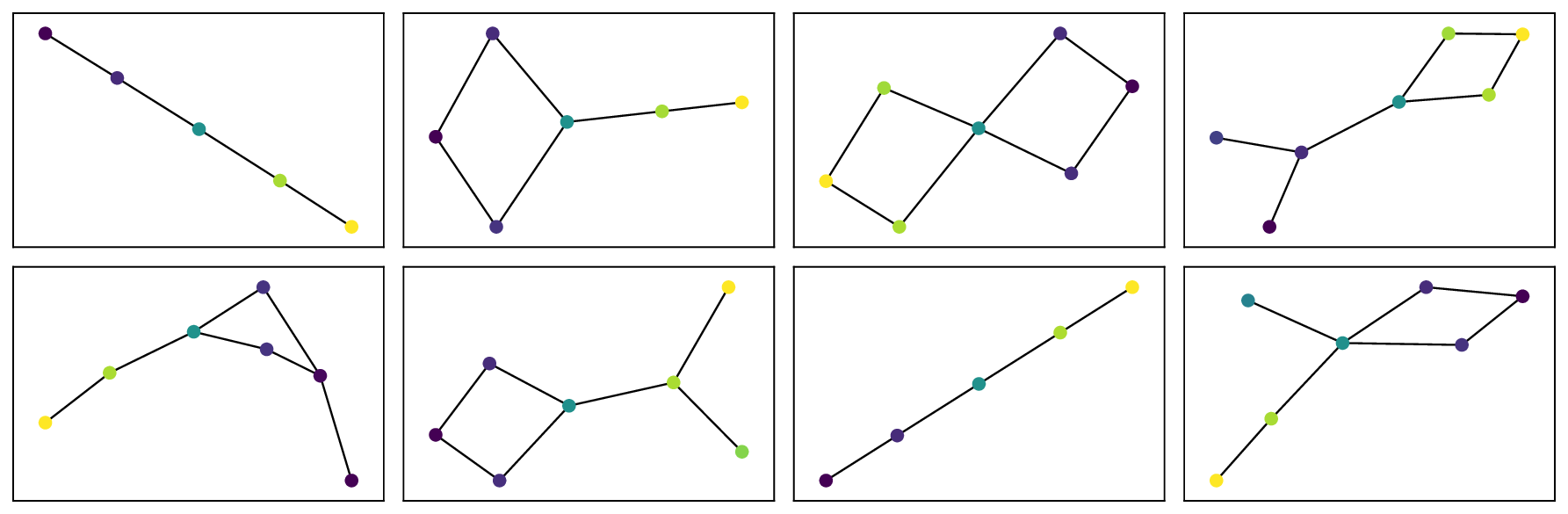}
\vspace{1mm}
\caption{Eight samples from the optimized GMM soft Mapper of the two intersecting circles dataset.}
\label{Samples_two_insert_cir}
\end{figure}

\begin{figure}[H]
\centering
\includegraphics[width=0.9\textwidth]{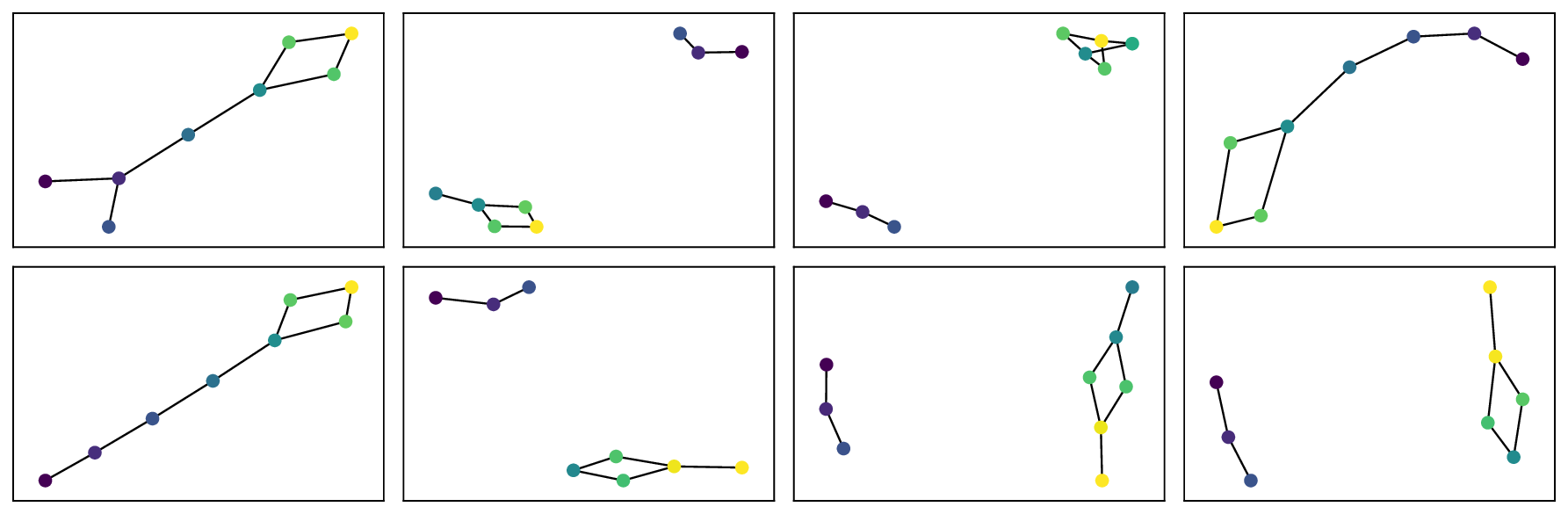}
\vspace{1mm}
\caption{Eight samples from the optimized GMM soft Mapper of the two disjoint \textcolor{black}{circles with different radii} dataset.}
\label{Samples_two_cir_extra}
\end{figure}

\begin{figure}[H]
\centering
\includegraphics[width=0.9\textwidth]{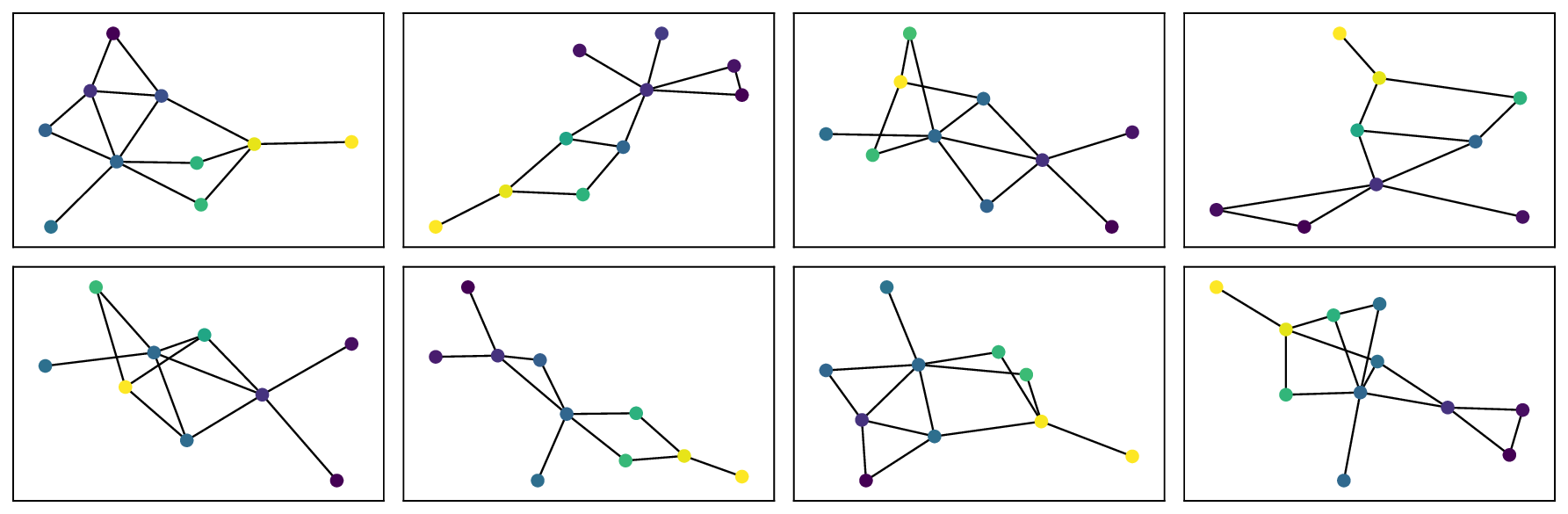}
\vspace{1mm}
\caption{Eight samples from the optimized GMM soft Mapper of the two intersecting \textcolor{black}{circles with different radii} dataset.}
\label{Samples_two_insert_cir_extra}
\end{figure}

\begin{figure}[H]
\centering
\includegraphics[width=0.9\textwidth]{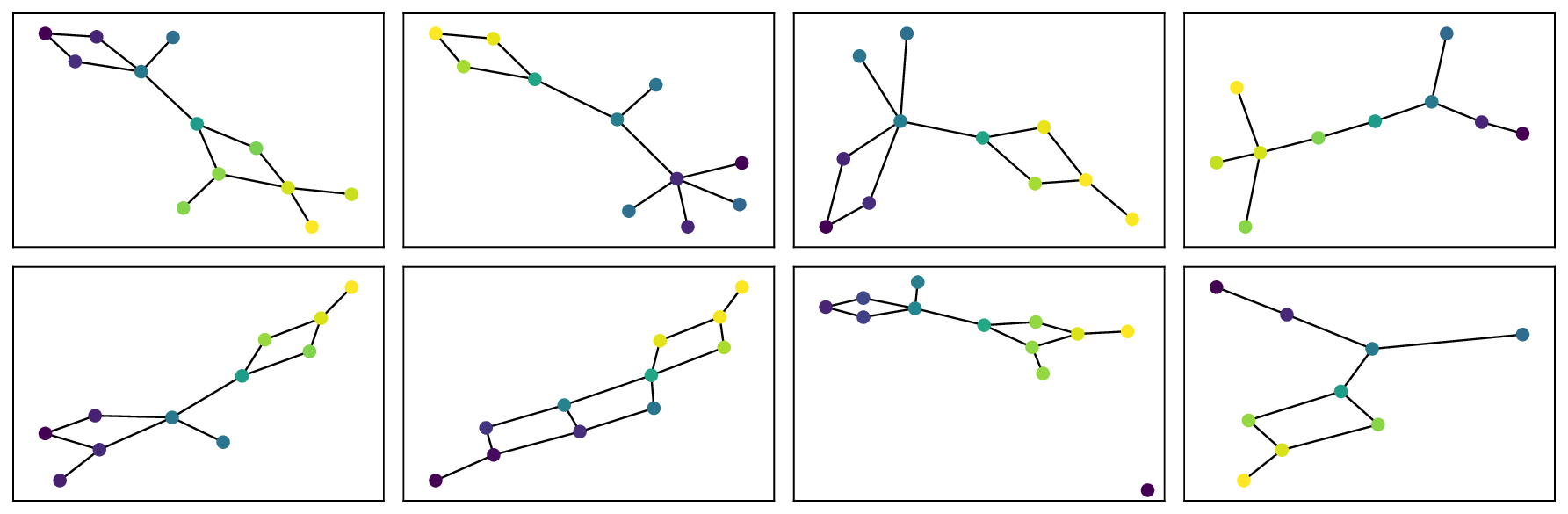}
\vspace{1mm}
\caption{Eight samples from the optimized GMM soft Mapper of the two circles with small noise dataset.}
\label{Samples_two_cir_noise}
\end{figure}

\begin{figure}[H]
\centering
\includegraphics[width=0.9\textwidth]{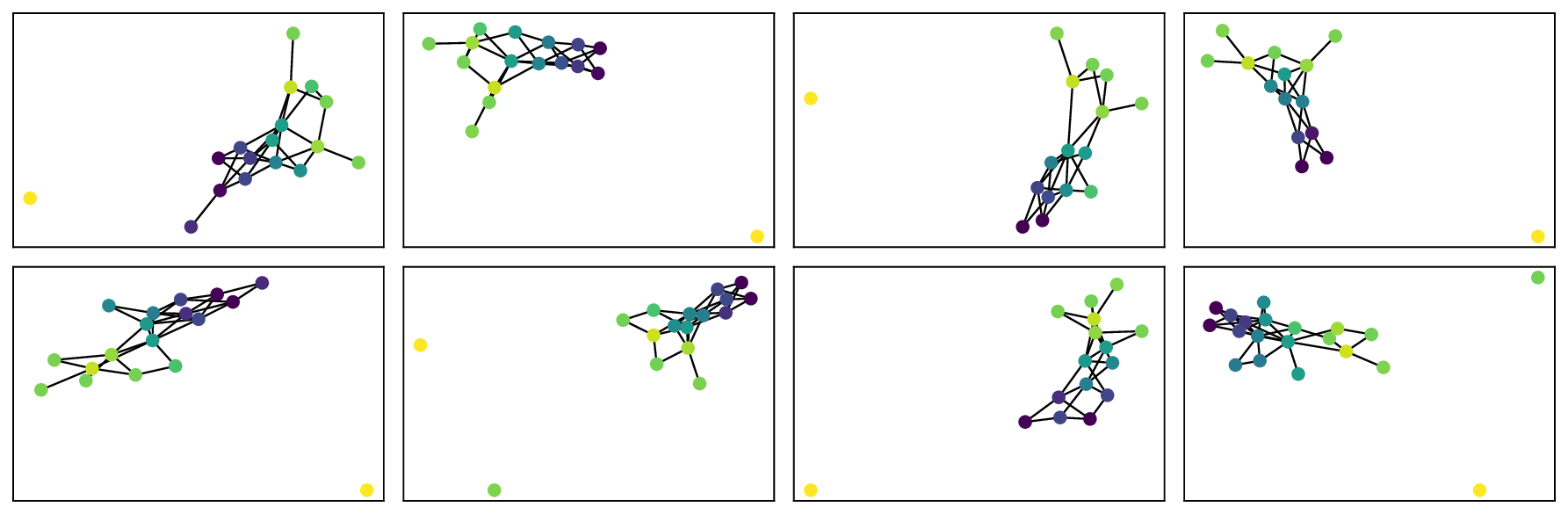}
\vspace{1mm}
\caption{Eight samples from the optimized GMM soft Mapper of the two intersecting circles with big noise dataset.}
\label{Samples_two_insert_cir_noise}
\end{figure}

\begin{figure}[H]
\centering
\includegraphics[width=0.9\textwidth]{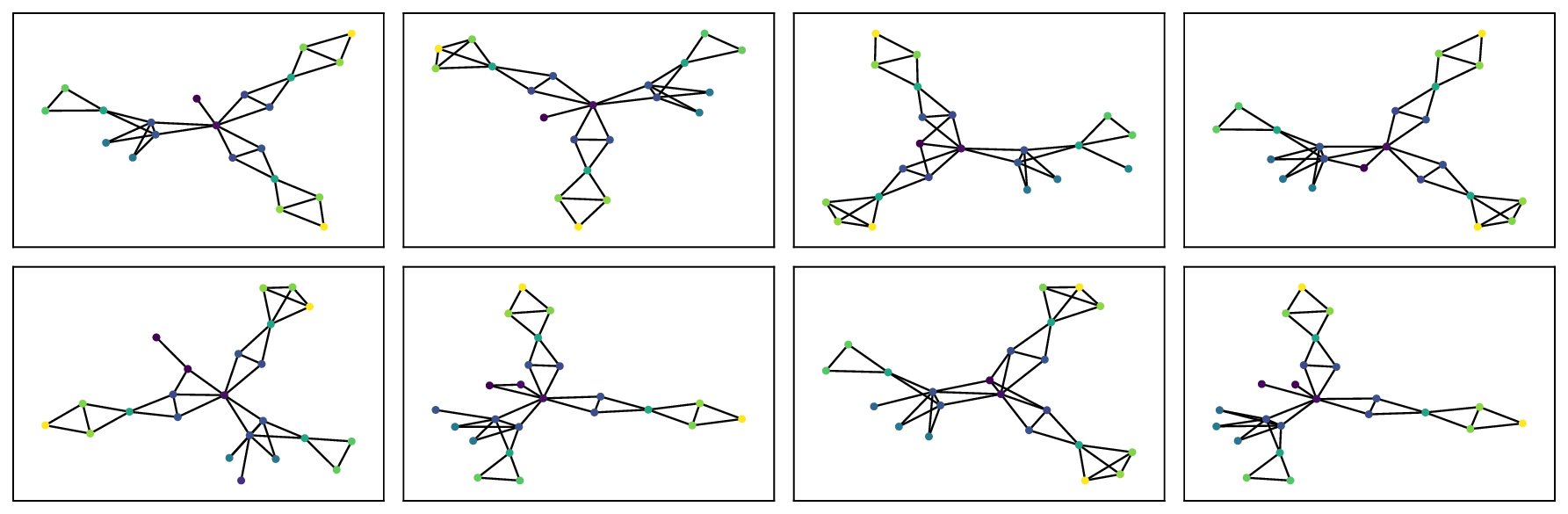}
\vspace{1mm}
\caption{Eight samples from the optimized GMM soft Mapper on the 3D human dataset.}
\label{Samples_3Dhuman}
\end{figure}

\begin{table}[htbp]
\centering
\caption{Statistical analysis of samples from GMM soft Mapper. TC is short for two circles. TIC is short for two intersecting circles.}
\label{combined_graph_results}
\small
\begin{tabular}{llrrrr}
\toprule
\textbf{Configuration} & \textbf{Metric} & \textbf{Mean} & \textbf{95\% CI} & \textbf{Median} & \textbf{Mode} \\
\midrule
\multirow{2}{*}{TC} 
    & Avg Degrees           & 1.98  & [1.97, 1.99] & 2.00 & 2.00 \\
    & Connected Components  & 1.98  & [1.97, 1.99] & 2.00 & 2.00 \\
    & Loops                 & 2.29  & [2.22, 2.35] & 2.00 & 2.00 \\
\midrule
\multirow{2}{*}{TIC} 
    & Avg Degrees           & 2.00  & [1.99, 2.01] & 2.00  & 2.00\\
    & Connected Components  & 1.01  & [1.00, 1.01] & 1.00 & 1.00\\
    & Loops                 & 1.15  & [1.11, 1.20] & 1.0  & 1.00 \\
\midrule
\multirow{2}{*}{3D Human} 
    & Avg Degrees           & 3.27  & [3.22, 3.23] & 3.22 & 3.25 \\
    & Connected Components  & 1.00  & [1.00, 1.01] & 1.00 & 1.00 \\
    & Loops                 & 94.22 & [89.93, 98.50] & 70.0 & 58.0 \\
\midrule
\multirow{2}{*}{Unequal TC} 
    & Avg Degrees           & 1.64  & [1.63, 1.65] & 1.71 & 1.75 \\
    & Connected Components  & 1.91  & [1.90, 1.93] & 2.00 & 2.00 \\
    & Loops                 & 0.75  & [0.71, 0.78] & 1.00 & 1.00 \\
\midrule
\multirow{2}{*}{Unequal TIC} 
    & Avg Degrees           & 2.62  & [2.61, 2.63] & 2.60 & 2.67 \\
    & Connected Components  & 1.01  & [1.00, 1.01] & 1.00 & 1.00 \\
    & Loops                 & 11.22 & [10.80, 11.63] & 10.00 & 10.0 \\
\midrule
\multirow{2}{*}{Small Noise} 
    & Avg Degrees           & 1.96  & [1.95, 1.97] & 2.00 & 2.00 \\
    & Connected Components  & 1.18  & [1.16, 1.20] & 1.00 & 1.00 \\
    & Loops                 & 1.27  & [1.20, 1.35] & 1.00 & 1.00 \\
\midrule
\multirow{2}{*}{Big Noise} 
    & Avg Degrees           & 3.27  & [3.25, 3.28] & 3.25 & 3.33 \\
    & Connected Components  & 2.37  & [2.34, 2.40] & 2.00 & 2.00 \\
    & Loops                 & 1296.32 & [1216.40, 1376.24] & 790.50 & 120 \\
\bottomrule
\end{tabular}
\end{table}

\newpage 

\section*{Appendix C. More explanations for the D-Mapper*}

\begin{figure}
\centering
\includegraphics[width=.8\textwidth]{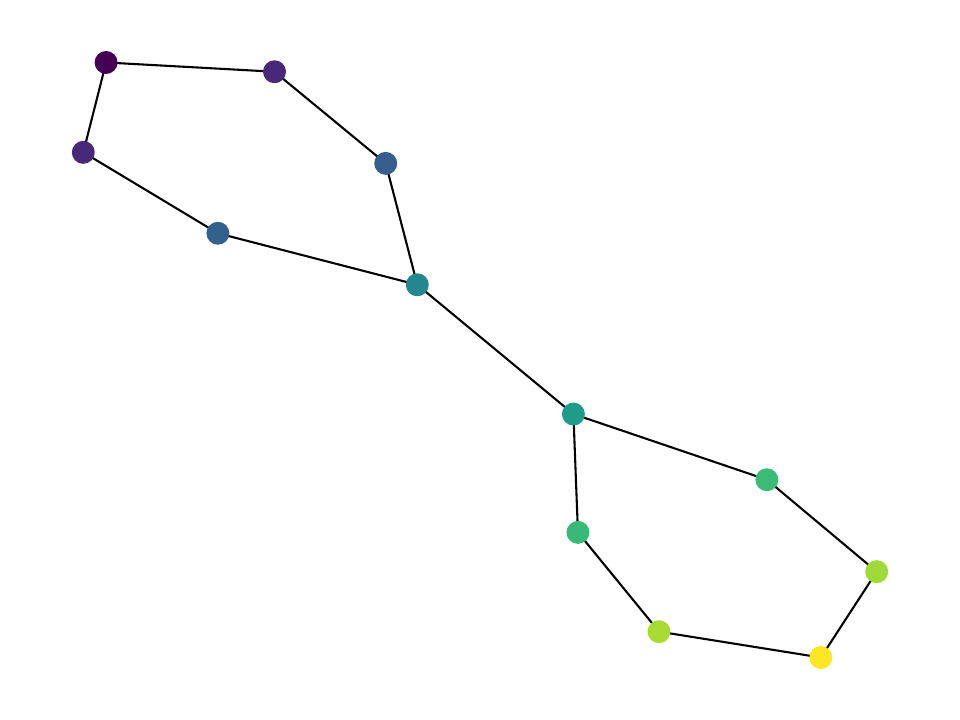}
\vspace{1mm}
\caption{D-Mapper graph of two intersecting circles dataset with number of intervals $K=8$.}
\label{D-Mapper_extra}
\end{figure}

For the two intersecting circles dataset with the number of   intervals $K = 8$, the results are shown in Figure \ref{D-Mapper_extra}. The experiment shows that the D-Mapper achieves a higher $TSR$ with a larger $K$, it is likely because that the increased number of groups in a  GMM leads to more fine components. However, we should be aware that a large number of intervals may also lead to spurious structures in the Mapper graph.

\section*{Appendix D. Supplementary figures for MSBB dataset of the brain area 36}
\begin{figure}[H]
\centering
  \begin{minipage}{0.45\textwidth}
    \centering
    \includegraphics[width=\linewidth]{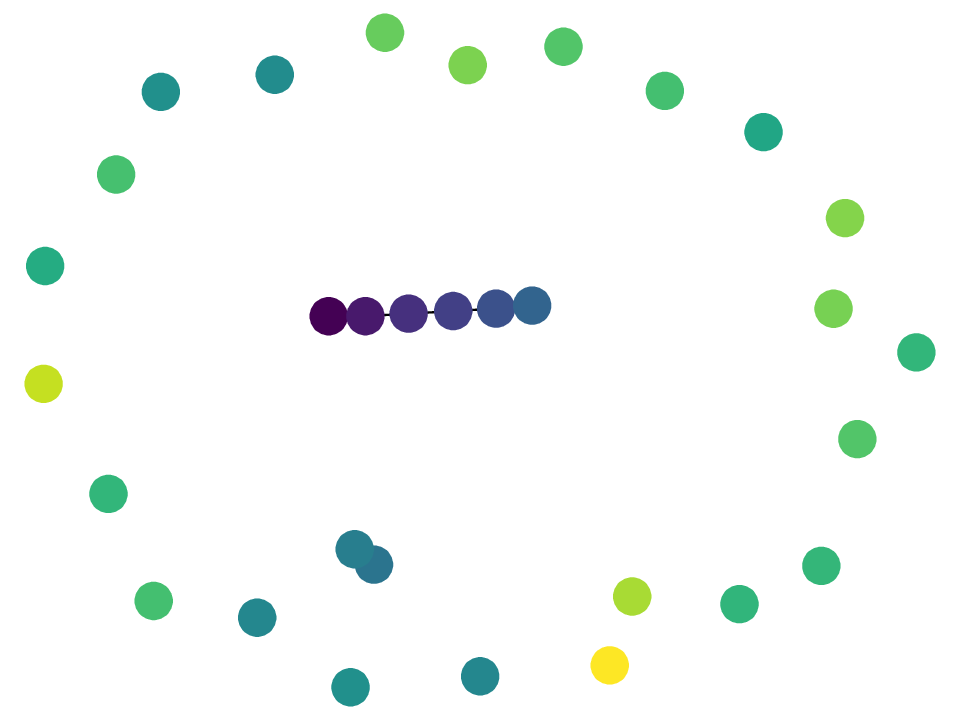}
    \caption{The graph mode of the GMM soft Mapper without optimization for the MSBB dataset of the brain area 36.}
  \end{minipage}%
  \hfill
  \begin{minipage}{0.45\textwidth}
    \centering
    \includegraphics[width=\linewidth]{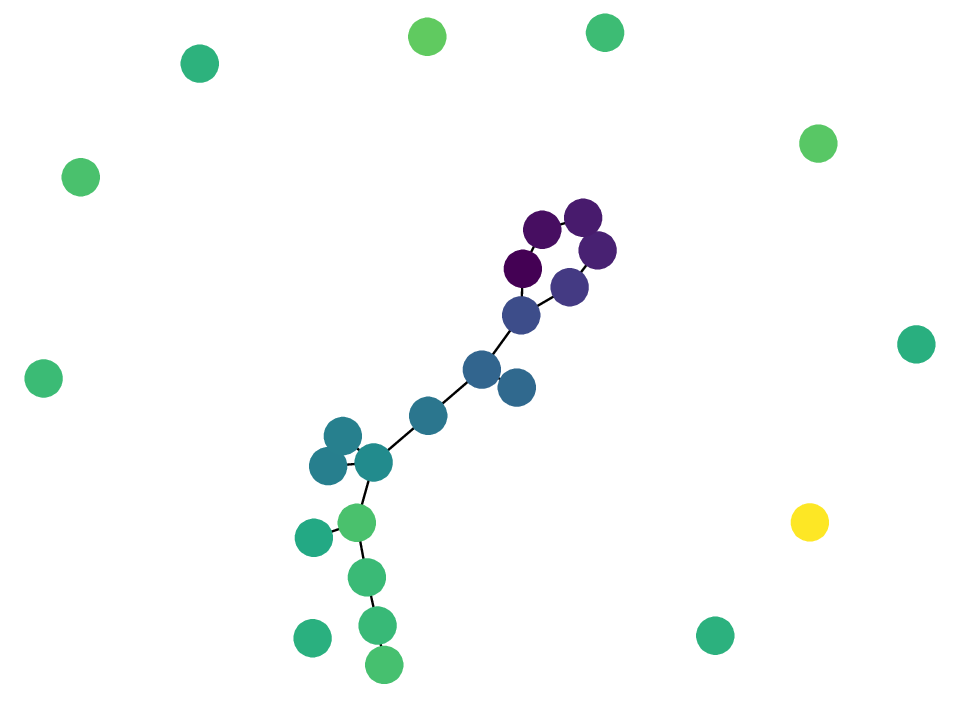}
    \caption{The graph mode of the GMM soft Mapper with optimization for the MSBB dataset of the brain area 36.}
  \end{minipage}
\end{figure}

\begin{figure}[H]
\centering
\includegraphics[width=1\textwidth]{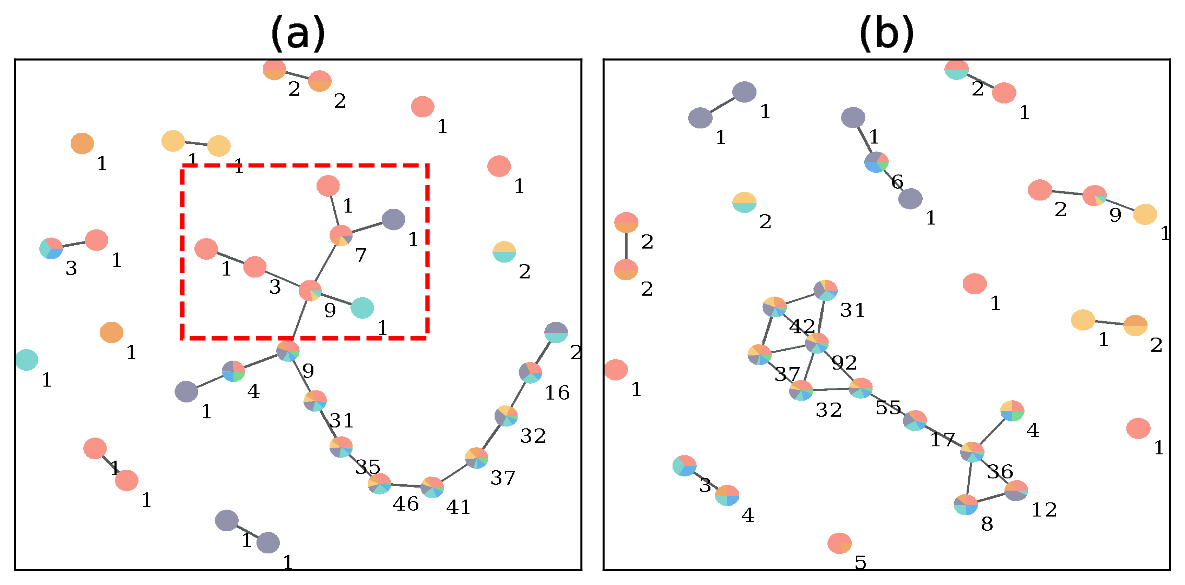}
\vspace{1mm}
\caption{(a) The standard Mapper graph for the MSBB dataset of the brain area 36. (b) The D-Mapper graph for the MSBB dataset of the brain area 36.}
\label{MSBB_compare}
\end{figure}

\section*{Appendix E. Loss figures}
\begin{figure}[H]
  \centering
  \begin{subfigure}[b]{0.24\textwidth}
    \includegraphics[width=\textwidth]{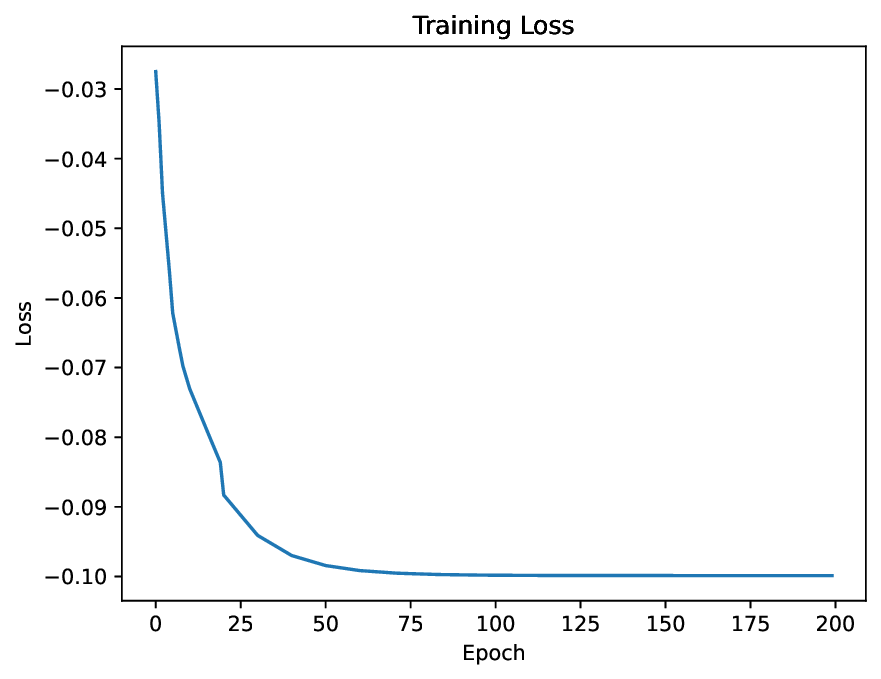}
    \caption{}
  \end{subfigure}
  \hfill
  \begin{subfigure}[b]{0.24\textwidth}
    \includegraphics[width=\textwidth]{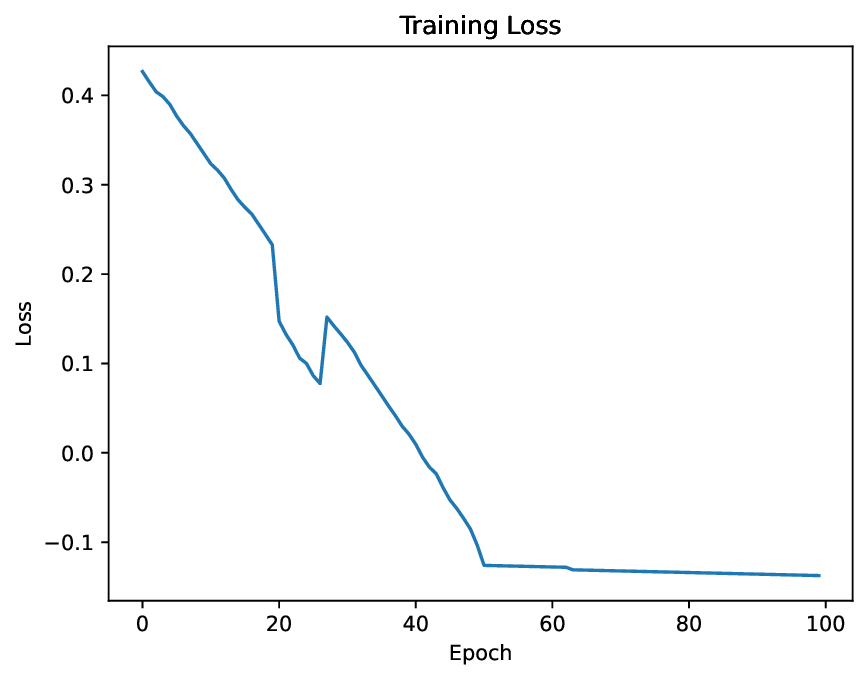}
    \caption{}
  \end{subfigure}
  \hfill
  \begin{subfigure}[b]{0.24\textwidth}
    \includegraphics[width=\textwidth]{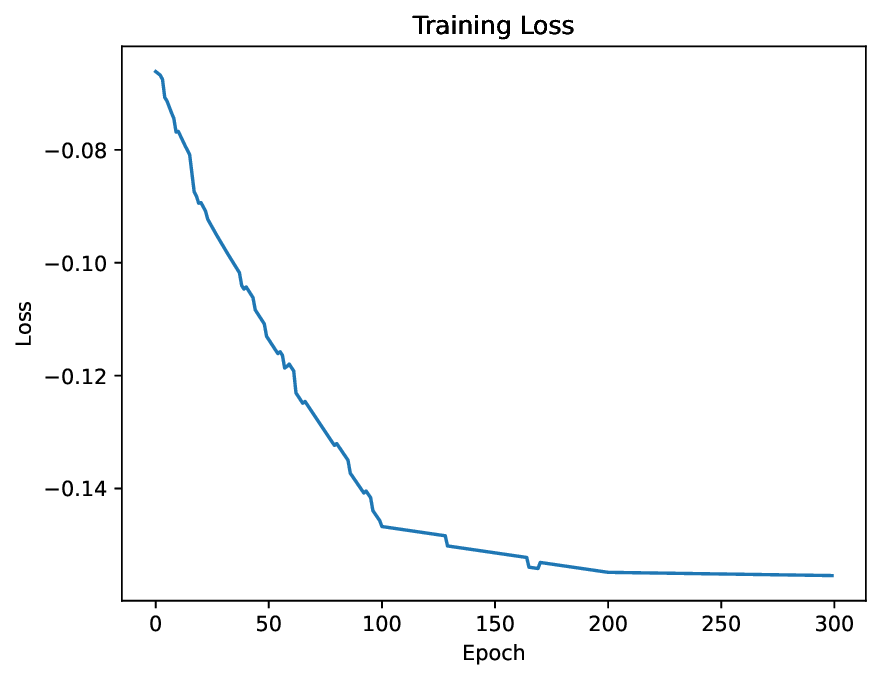}
    \caption{}
  \end{subfigure}
  \hfill
  \begin{subfigure}[b]{0.24\textwidth}
    \includegraphics[width=\textwidth]{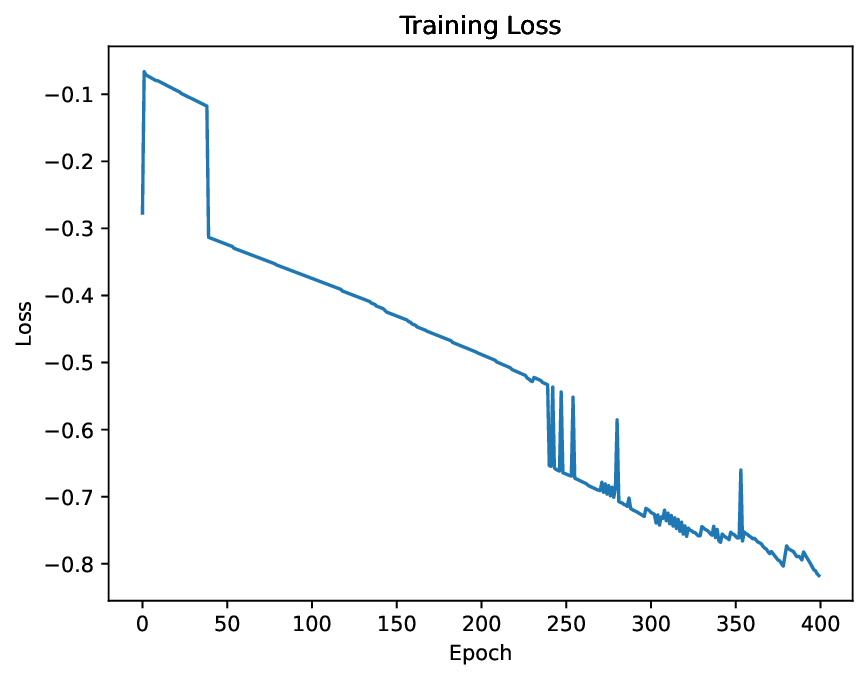}
    \caption{}
  \end{subfigure}
  
  \begin{subfigure}[b]{0.24\textwidth}
    \includegraphics[width=\textwidth]{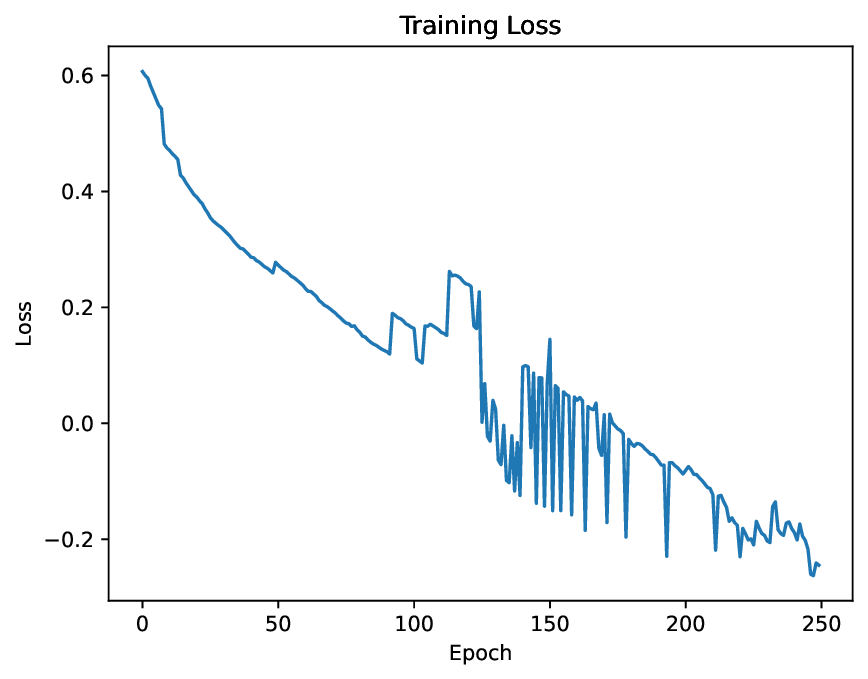}
    \caption{}
  \end{subfigure}
  \hfill
  \begin{subfigure}[b]{0.24\textwidth}
    \includegraphics[width=\textwidth]{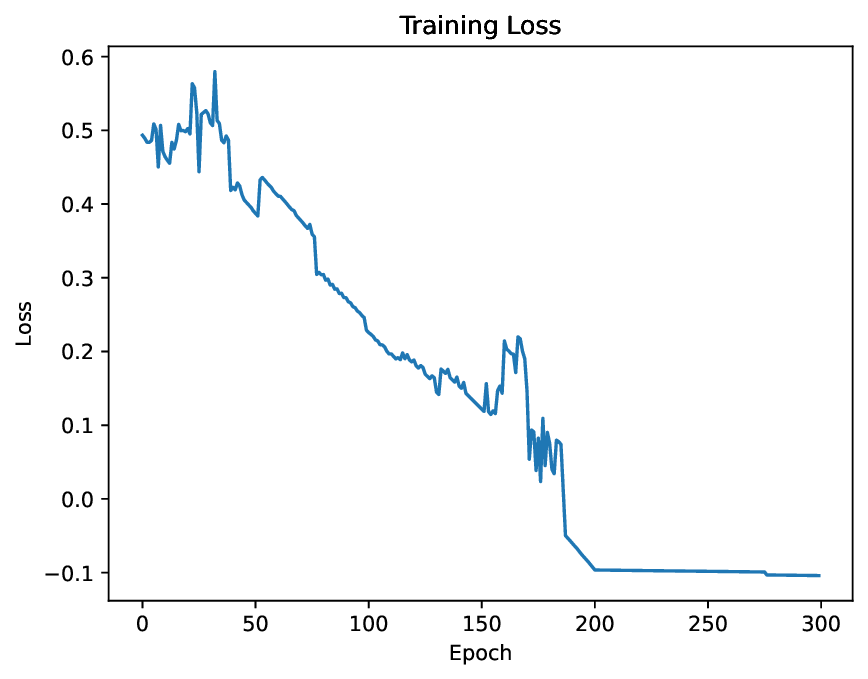}
    \caption{}
  \end{subfigure}
  \hfill
  \begin{subfigure}[b]{0.24\textwidth}
    \includegraphics[width=\textwidth]{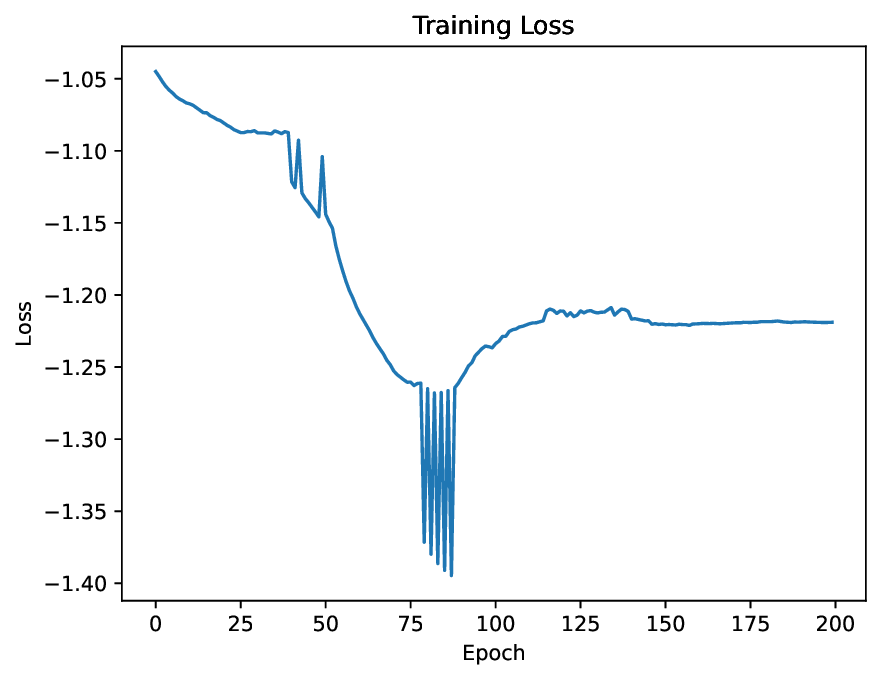}
    \caption{}
  \end{subfigure}
  \hfill
  \begin{subfigure}[b]{0.24\textwidth}
    \includegraphics[width=\textwidth]{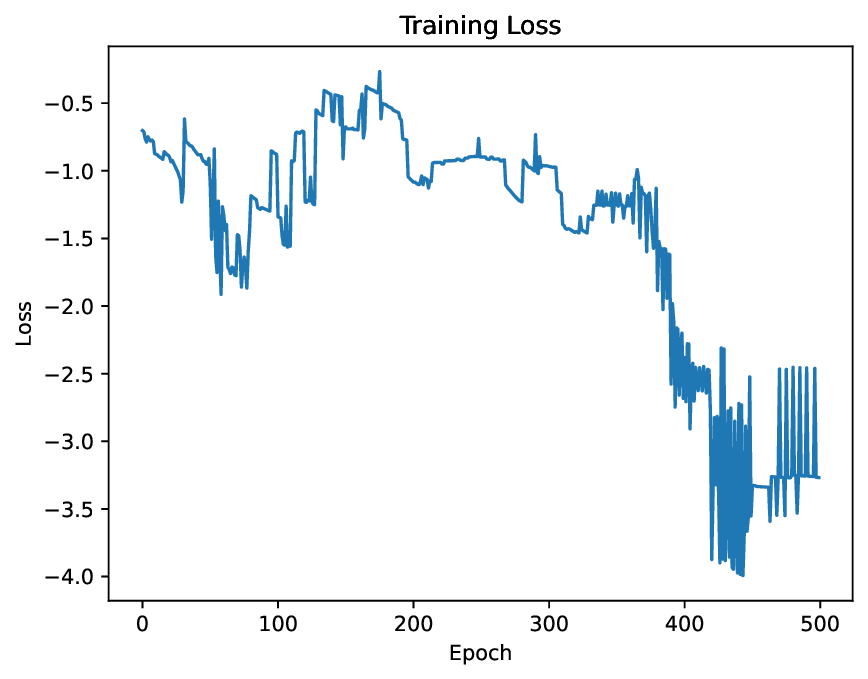}
    \caption{}
  \end{subfigure}
  
  \caption{Training losses of the GMM soft Mapper while optimizing prameters across different datasets. (a)  The dataset of the two disjoint circles. (b)  The dataset of the two intersecting circles dataset. (c)  The dataset of the \textcolor{black}{circles with different radii} dataset. (d) The dataset of the two intersecting \textcolor{black}{circles with different radii} dataset. (e) The dataset of the two intersecting circles with small noises. (f) The dataset of the two intersecting circles with large noises. (g) The 3D human dataset. (h) The MSBB dataset of the Brain area 36.}
\end{figure}
\end{document}